\documentclass[journal]{IEEEtran}

\usepackage{caption}
\usepackage{graphicx}
\usepackage{booktabs}
\usepackage{subfigure}
\usepackage{cite}
\usepackage{amsmath,amssymb,amsfonts}
\usepackage{algorithmic}
\usepackage{textcomp}
\usepackage{lineno}
\usepackage{times}
\usepackage{epsfig}

\usepackage{booktabs}
\usepackage{epstopdf}

\usepackage{subeqnarray}
\usepackage{color}
\usepackage{amstext}
\usepackage{verbatim}
\usepackage{array}
\usepackage{boldline}
\usepackage{makecell}

\usepackage{setspace}
\usepackage{threeparttable}

\usepackage{chngpage}
\usepackage{multirow}
\usepackage{chngpage}
\usepackage{setspace}
\usepackage{colortbl}

\definecolor{lightgray}{gray}{.92}
\definecolor{tinygray}{gray}{.96}
\definecolor{mycolor}{RGB}{255,0,0}

\graphicspath{{figures/},{figure1/},{figure2/},{figure3/}}

\usepackage[bookmarks,colorlinks,linkcolor=red, anchorcolor=blue, citecolor=green]{hyperref}
\usepackage[capitalize]{cleveref}
\begin{document}

\title{Robust Deep Ensemble Method for Real-world Image Denoising}

\author{
        Pengju~Liu,~
        Hongzhi~Zhang,~
        Jinghui~Wang,~        Yuzhi~Wang~
        Dongwei~Ren, 
        and Wangmeng~Zuo
\thanks{P. Liu, H. Zhang, D. Ren, and W. Zuo are with the School of Computer Science and Technology, Harbin Institute of Technology, Harbin, China.
        e-mail: lpj008@126.com; zhanghz0451@gmail.com; csdren@hit.edu.cn; wmzuo@hit.edu.cn.} 
\thanks{J. Wang and Y. Wang are with Megvii Inc, Beijing, China. e-mail: wangjinghui@megvii.com; wangyuzhi@megvii.com.}
}%

\maketitle

\begin{abstract}
	
	Recently, deep learning-based image denoising methods have achieved promising performance on test data with the same distribution as training set, where various denoising models based on synthetic or collected real-world training data have been learned. 
	However, when handling real-world noisy images, the denoising performance is still limited.  
	In this paper, we propose a simple yet effective Bayesian deep ensemble (BDE) method for real-world image denoising, where several representative deep denoisers pre-trained with various training data settings can be fused to improve robustness for real-world noisy images. 
	The foundation of BDE is that real-world image noises are highly signal-dependent, and heterogeneous noises in a real-world noisy image can be separately handled by different denoisers.  
	In particular, we take well-trained CBDNet, NBNet, HINet, Uformer and GMSNet into denoiser pool, and a U-Net is adopted to predict pixel-wise weighting maps to fuse these denoisers. 
	Instead of solely learning pixel-wise weighting maps, Bayesian deep learning strategy is introduced to predict weighting uncertainty as well as weighting map, by which prediction variance can be modeled for improving robustness on real-world noisy images. 
	Extensive experiments have shown that real-world noises can be better removed by fusing existing denoisers instead of training a big denoiser with expensive cost.  
	On DND dataset, our BDE achieves +0.28~dB PSNR gain over the state-of-the-art denoising method. 
	Moreover, we note that our BDE denoiser based on different Gaussian noise levels outperforms state-of-the-art CBDNet when applying to real-world noisy images. 
	Furthermore, our BDE can be extended to other image restoration tasks, including image deblurring, image deraining and single image super-resolution. 
	On benchmark datasets, our method achieves +0.30dB, +0.18dB and +0.12dB PSNR gains for image deblurring, image deraining and single image super-resolution, respectively.

\end{abstract}
 

\IEEEpeerreviewmaketitle

\section{Introduction}
\begin{figure}[!t]
  \vspace{-0ex}
  \begin{center}
    \vspace{-0.0ex}
    \includegraphics[width=0.475\textwidth]{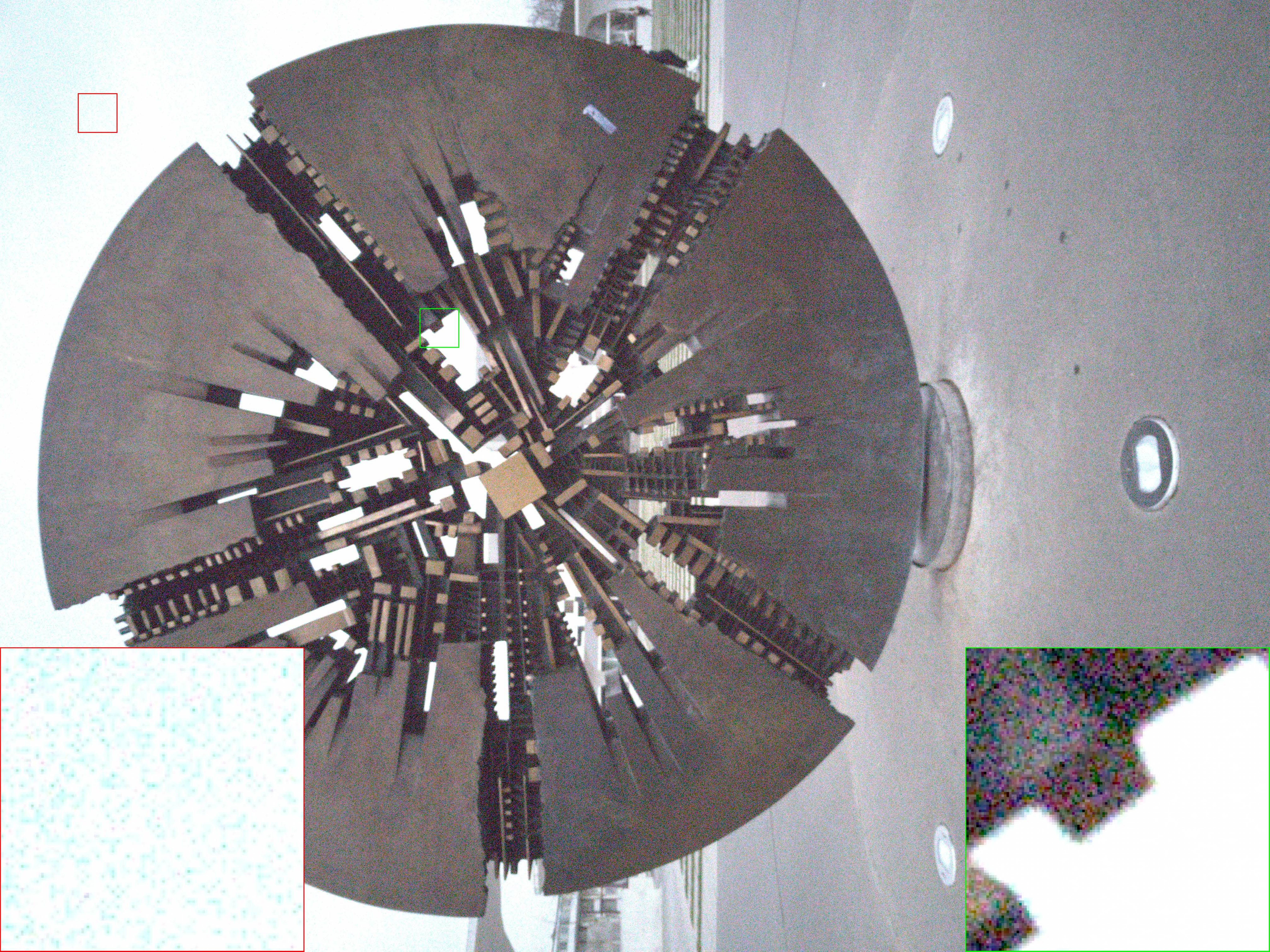}
    \vspace{-0.0ex}
  \end{center}
    \caption{Heterogeneity distribution of noises in real-world noisy image. 
    }
    \vspace{-0ex}
    \label{fig:noisyimage}
\end{figure}

Image denoising, aiming to recover latent clean images from noisy images, is a fundamental task in image processing and low-level computer vision field.
In recent years, deep learning-based denoising methods flourish in this field and achieve state-of-the-art denoising performance. 
Especially, for additive-white-Gaussian-noise (AWGN) removal, a basic testing bed, convolutional neural network (CNN) is successfully developed for achieving superior denoising performance \cite{zhang2017beyond} over traditional denoising methods\cite{dabov2007image,nasri2009image,zhang2014adaptive,li2016image}.
Subsequently, various CNN-based denoisers ~\cite{mao2016image,tai2017memnet,zhang2018ffdnet,liu2018multi,peng2019dilated,zhang2020residual,liu2018non,wang2021deep,cheng2021nbnet,Zheng_2021_CVPR} have been studied for boosting quantitative AWGN denoising performance.
However, these CNN-based Gaussian denoisers usually perform limited, when handling real-world noisy images, due to the distribution gap between Gaussian noises and real-world noises. 
%
%
Moreover, as shown in Fig. \ref{fig:noisyimage}, noises in real-world noisy image are signal-dependent, and it is more difficult to remove heterogeneous noises for one denoiser. 
In~\cite{zhang2018ffdnet}, FFDNet is suggested to adopt different noise reduction intensities for region-aware denoising.
However, FFDNet needs to provide spatially variant noise map, and is still developed for Gaussian denoising. 
  
To remove real-world noises, supervised denoising methods can be categorized into two groups in terms of training pairs, i.e., synthesizing realistic noisy images
\cite{guo2019toward,chen2018learning,brooks2019unprocessing,wang2020practical,zhu2016noise,liu2007automatic} and collecting real-world noisy/clean pairs~\cite{SIDD_2018_CVPR}.
These methods have achieved promising performance on test data with the same distribution as the training set. 
  However, they usually suffer from notable performance declines when handling real-world noisy images with unknown noise distribution.
  %
  In~\cite{song2020grouped}, GMSNet is trained on both real-world and synthetic noisy/clean pairs to obtain better generalization performance.
  Nonetheless, its denoising performance is still limited for real-world noisy images with heterogeneous noise distribution. 
  As a summary, it is hard for a single denoiser to obtain satisfying denoising performance for spatially heterogeneous noisy images. 
  
  \begin{figure*}[!htbp]
	\vspace{-0ex}
	\centering
    \subfigure[Noisy Image]{
      \begin{minipage}[c]{0.24\textwidth}
      \centering
        \includegraphics[width=0.99\linewidth]{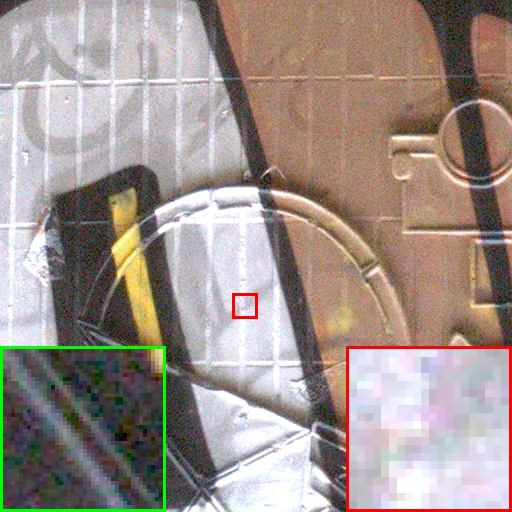}
        \label{fig:noisya}
      \end{minipage}%
      }
  \hspace{-1ex}
  \subfigure[CBDNet~\cite{guo2019toward} 33.62]{
    \begin{minipage}[c]{0.24\textwidth}
    \centering
      \includegraphics[width=0.99\linewidth]{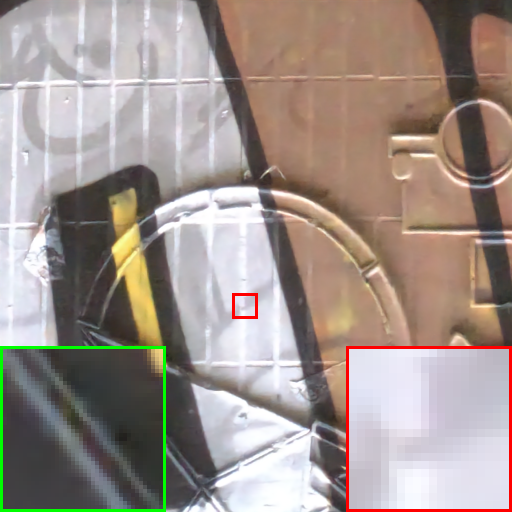}
      \label{fig:short-CBDNet}
    \end{minipage}%
    }
	\hspace{-1ex}
    \subfigure[GMSNet~\cite{song2020grouped} 34.70]{
      \begin{minipage}[c]{0.24\textwidth}
      \centering
        \includegraphics[width=0.99\linewidth]{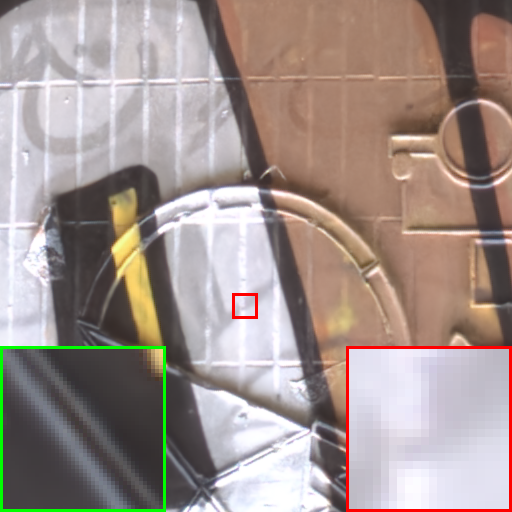}
        \label{fig:short-GMSNet}
      \end{minipage}%
      }
	\hspace{-1ex}
  \subfigure[BDE 34.94]{
    \begin{minipage}[c]{0.24\textwidth}
    \centering
      \includegraphics[width=0.99\linewidth]{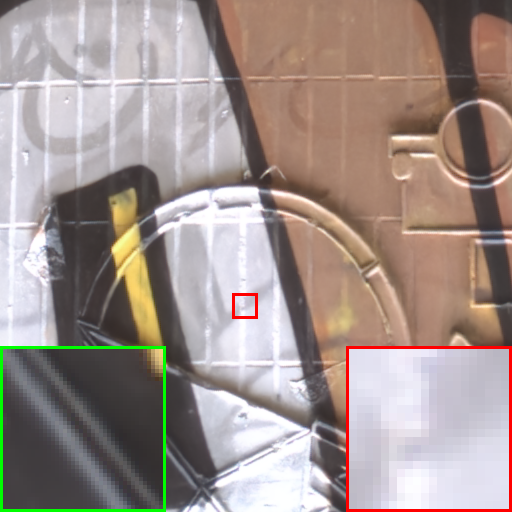}
      \label{fig:short-UEM}
    \end{minipage}%
    }
	\vspace{-0ex}
	\caption{ An example of fusing CBDNet~\cite{guo2019toward} and GMSNet~\cite{song2020grouped} by our BDE.
	  It can be easily found that our BDE is a region-aware method to fuse different denoisers: 
	  for the region in the green box, our BDE mainly comes from GMSNet~\cite{song2020grouped}, while in the red box, our BDE can be seen as the fusion of CBDNet~\cite{guo2019toward} and GMSNet~\cite{song2020grouped}. 
	  }
	  \label{fig:sel}
	  \vspace{-0ex}
\end{figure*}
  
  In this paper, we propose a simple yet effective Bayesian deep ensemble (BDE) method for real-world image denoising, where several representative deep denoisers pre-trained with various training data settings can be fused to improve robustness for real-world noisy images. 
  The foundation of BDE is that real-world image noises are highly signal-dependent, and heterogeneous noises in a real-world noisy image can be separately handled by different denoisers.  
  In particular, we take well-trained CBDNet~\cite{guo2019toward}, NBNet~\cite{cheng2021nbnet}, HINet~\cite{chen2021hinet}, Uformer~\cite{wang2021uformer}, and GMSNet~\cite{song2020grouped} as our denoiser pool, and a U-Net is adopted to predict pixel-wise weighting maps to fuse these denoisers.   
  Instead of solely learning pixel-wise weighting maps, Bayesian deep learning strategy is introduced to predict weighting uncertainty as well as weighting map, by which prediction variance can be modeled for improving robustness on real-world noisy images. 
  As shown in Fig. \ref{fig:sel}, our BDE outperforms individual state-of-the-art denoisers. 

  Extensive experiments have shown that real-world noises can be better removed by fusing existing denoisers instead of training a big denoiser with expensive cost.  
  We show the effectiveness and efficiency of our method on two real-world benchmarks, \emph{i.e.} Darmstadt noise dataset (DND)~\cite{plotz2017benchmarking} and smartphone image denoising dataset (SIDD)~\cite{SIDD_2018_CVPR}.
  Note that the DND benchmark is taken as the main benchmark, due to none of the corresponding training set and ground truth of testing images.
  On DND dataset, our BDE achieves +0.28~dB PSNR gain over the state-of-the-art denoising method. 
  
  One may notice that there is an ensemble method for image denoising  \cite{choi2019optimal}, where CSNet is proposed to combine five denoisers with different noise levels. 
  Our BDE is distinctive with CSNet from two perspectives:
  (i) CSNet is developed for gray-style Gaussian denoising task, while our BDE aims to fuse denoisers for removing real-world noises. 
  (ii) CSNet adopts an image-level ensemble strategy, by which pre-trained denoisers cannot be well exploited and fused. 
  In our BDE, pixel-wise ensemble strategy benefits heterogeneous noise distribution. 
  As a support, BDE denoiser based on different Gaussian noise levels outperforms state-of-the-art CBDNet when applying to real-world noisy images.

  Furthermore, our BDE can be extended to other image restoration tasks, including image deblurring, image deraining and single image super-resolution, where the proposed ensemble method can be directly applied to fuse representative restoration models for each task.
  On benchmark datasets, our method achieves +0.30dB, +0.18dB and +0.12dB PSNR gains for image deblurring, image deraining and single image super-resolution, respectively. 

  The contribution of this paper can be summarized from three perspectives.
  \begin {itemize}
	%
	 \item  
	 A robust ensemble method is proposed to handle heterogeneity distribution in real-world noisy images, where well-trained denoisers can be fused in a pixel-wise fusion manner. 
	 \item
	 Bayesian deep ensemble method is adopted to predict weighting uncertainty as well as weighting map, by which prediction variance can be modeled for improving denoising robustness. 
	 On both DND and SIDD benchmarks, our method achieves better results than state-of-the-art denoisers. 
	\item 
	Our ensemble method can be easily extended to other restoration tasks including image deblurring, image deraining and single image super-resolution, where notable gains are achieved on 14 benchmark datasets. 
\end {itemize}

The remainder of this paper is organized as follows: In Section~\ref{secII}, we review image denoising and ensemble deep learning methods. 
In Section~\ref{secIII}, we give a detailed introduction of our robust ensemble method for image denoising, as well as its extension to other image restoration tasks. 
In Section~\ref{secIV}, we conduct experiments to evaluate our method on both real-world noisy benchmarks and other image restoration tasks, in comparison to corresponding state-of-the-art methods. 
Finally, some conclusion remarks are given in Section~\ref{secV}.

\section{Related Works}\label{secII}
%

\subsection{Image denoising}
Nowadays, most image denoising methods focus on CNN-based methods since DnCNN~\cite{zhang2017beyond}, which achieve better performance than traditional methods~\cite{dabov2007image}.
According to the training set, we mainly review two types of denoisers, \emph{i.e.} training on the synthetic dataset and real-world dataset.

\subsubsection{Image Denoising with Synthetic Dataset}

The CNN-based image denoising can be dated back to 2009~\cite{jain2008natural}, however, its performance could only compare to BM3D~\cite{dabov2007image}.
Unfortunately, the successive denoisers in \cite{xie2012image,burger2012image} also do not achieve success.
The milestone of image denoising is DnCNN~\cite{zhang2017beyond}, which combines residual learning stage and batch normalization (BN)~\cite{ioffe2015batch} and achieves the state-of-the-art performance.
Subsequently, some methods, such as REDNet~\cite{mao2016image}, MemNet~\cite{tai2017memnet}, MWCNN~\cite{liu2018multi}, DSNet~\cite{peng2019dilated}, RDN~\cite{zhang2020residual}, and DCDicL~\cite{Zheng_2021_CVPR}  propose a more complex architecture to obtain performance progress.
Instead of designing the network architecture, some methods utilize some image priors, such as non-local information and sparse representation, to promote the denoising performance. 
In N3Net~\cite{plotz2018neural} and NLRN~\cite{liu2018non}, they exploit the non-local image prior by proposing non-local modules for noise removal.
While Cheng \emph{et al.} \cite{cheng2021nbnet} propose a non-local attention module to explicitly learn the basis generation as well as subspace projection.
Wang \emph{et al.} \cite{wang2021deep} incorporate the global non-linear smoothness constraint prior to CNN which can improve the interpretability of deep learning.
To train controllable denoiser, \cite{zhang2018ffdnet,du2020flexible,zhang2018learning} take the noisy image with its noise map as input.
However, those methods are trained and tested on Gaussian noise, and most of them are less effectiveness even useless when they are applied to real-world noisy images due to the huge domain gap.

Referring to the camera image signal processing (ISP) pipeline, more complex noise can be synthesized for removing the real-world noisy images~\cite{guo2019toward,zamir2020cycleisp}.
Some methods~\cite{chen2018learning,brooks2019unprocessing,wang2020practical} even process the noisy images in the raw domain, which directly read out from the sensor in the raw Bayer format before demosaicing and other ISP-processing, for better denoising performance.
However, the denoising performance of those denoisers on a real noisy dataset relies heavily on the similarity between image priors and noise statistics.
While the domain gap between synthetic noise and real noise is still non-negligible.
For example, the denoising performance of CBDNet~\cite{guo2019toward} on DND and SIDD is much weaker than Uformer~\cite{wang2021uformer} which is trained on real-world dataset.

\subsubsection{Real-world denoisers}
To remove real-world noise, the simple and effective approach is collecting real-world noisy/clean pairs, such as SIDD~\cite{SIDD_2018_CVPR}.
Then numerous denoisers, such as MPRNet~\cite{mehri2021mprnet},  MIRNet~\cite{zamir2020learning},  NBNet~\cite{cheng2021nbnet},   Uformer~\cite{wang2021uformer}, and HINet~\cite{chen2021hinet}, are trained only SIDD dataset and achieve impressive denoising performance on both SIDD and DND benchmarks.
However, collecting real-world noisy/clean image pairs is extremely challenging and non-trivial.
Those denoisers also fall into overfitting the training set and lack generalization. 
Training on both synthetic dataset and real-world dataset can alleviate the overfitting problem~\cite{anwar2019real,zamir2020cycleisp,jang2020dual,song2020grouped}. 
GMSNet~\cite{song2020grouped} achieves impressive denoising performance on the DND benchmark, which has no corresponding training set.
Recently, the unsupervised and self-supervised learning are introduced for real-world image denoising because of no training pairs ~\cite{ulyanov2018deep,lehtinen2018noise2noise,krull2019noise2void,batson2019noise2self,quan2020self2self,moran2020noisier2noise,xu2020noisy,byun2021fbi,laine2019high,wu2020unpaired,huang2021neighbor2neighbor}. 
Those methods often design specific architecture, such as blind-spot mechanism~\cite{krull2019noise2void}, to learn self-information for the noisy image itself.
Naturally, the domain gap is neglected to obtain generalization.
However, the denoising performance of those denoisers is still far less than the supervised learning, and they are often very time-consuming.

\subsection{Bayesian deep learning}
In Bayesian deep learning, input-dependent aleatoric and epistemic uncertainty are two main types of uncertainty.
The input-dependent aleatoric uncertainty, which is also known as heteroscedastic uncertainty, measures the inherent noise in observation data and can't be reduced by providing more data. 
Monte Carlo dropout~\cite{gal2016dropout}, which utilizes the dropout layer before sending to the convolution layer, is often adopted to simulate aleatoric uncertainty. 
The epistemic uncertainty models the uncertainty in the model parameters due to limited data and knowledge.
To capture both of them,  Kendall \emph{et al.} \cite{kendall2017uncertainties} propose a unified framework and improve the robustness and performance on depth regression and semantic segmentation tasks. 
Recently, Bayesian deep learning is introduced to many vision tasks,  \emph{e.g.}  face recognition~\cite{chang2020data}, cross-task~\cite{zamir2020robust}, multi-source transfer learning~\cite{CHANDRA202054},  classification~\cite{zhao2020probabilistic}, and semantic segmentation~\cite{yeo2021robustness}, to improve the robustness of CNN models.

\subsection{Ensemble learning} \label{ensemble_model}
Our work is an ensemble deep learning method \cite{yeo2021robustness,kumar2020ensemble,hao2020visual,pinto2021ensemble,ganaie2021ensemble,zhang2021new}: it can combine several pretrained methods to obtain better generalization performance, naturally, the excavated prior can be well-used for the current task. 
In \cite{liu2020improved}, Liu \emph{et al.} adopt a spatial local fusion strategy to combining nonlocally centralized sparse representation (NCSR)~\cite{dong2012nonlocally} and FFDNet~\cite{zhang2018ffdnet} for improving Gaussian image denoising performance.
Yang \emph{et al.}~\cite{yang2020image} stack several base denoisers, \emph{i.e.} the well-established shrinkage (attenuation) based denoising method, for further performance improvement.
Choi \emph{et al.}~\cite{choi2019optimal} combine several Gaussian denoisers by estimating the mean squared error (MSE) of denoised images and then estimating combining weights with convex optimization.
The proposed method can well combine denoising results with accurately predicted MSE.
However, if the MSE is inaccurate, it may obtain a worse result. 

For other image restoration tasks, ensemble methods are also proposed for improving image restoration performance.
Liao \emph{et al.}~\cite{liao2015video} propose a two-stage video super-resolution ensemble method that the first and second stages aim at merging SR drafts and final deconvolution respectively.
In \cite{liu2017robust}, Liu \emph{et al.} propose a temporal adaptive neural network that takes as input a number of aligned LR frames and applies filters of different temporal
sizes to generate multiple HR frame estimates.
\cite{cho2021compression} proposes an ensemble method for compression artifacts reduction by fusing different restoration methods on a feature map level.

\section{Proposed Method}~\label{secIII}
In this section, we first present the motivation of our robust ensemble method, then the proposed Bayesian deep ensemble (BDE) denoising method is proposed, and finally BDE is extended to other image restoration tasks, including image deblurring, image deraining and single image super-resolution. 

\subsection{Problem Formulation and Motivation}
Given a noisy image $\mathbf{y}  $, the latent clean image $\mathbf{z}$ can be recovered by a denoiser $\mathcal{D}$
\begin{eqnarray} \label{eq:denoise1}
 \mathbf{\hat{z}} =   \mathcal{D} \left( \mathbf{y} \right).  
\end{eqnarray}
To train the denoiser $\mathcal{D}$, massive training pairs $\{\mathbf{y}_n, \mathbf{x}_n^{gt}\}_{n=1}^N$ should be prepared, where $\mathbf{x}_n^{gt}$ is the ground-truth clean image and $N$ is the number of training pairs. 
Then various loss functions, e.g., $\ell_1$-norm \cite{wang2018esrgan}, mean square error \cite{dong2015image,shi2016real,zhang2017beyond}, perceptual loss \cite{johnson2016perceptual} and adversarial loss \cite{ledig2017photo}, can be adopted to learn the parameters of denoiser $\mathcal{D}$ by closing the distance between denoised result $\hat{\mathbf{z}}_i$ and ground-truth clean image $\mathbf{x}_i^{gt}$. 
For real-world noisy images, training pairs can be obtained by synthesizing realistic noisy images
\cite{guo2019toward,chen2018learning,brooks2019unprocessing,wang2020practical,zhu2016noise,liu2007automatic} and collecting real-world noisy/clean pairs~\cite{SIDD_2018_CVPR}.
Recently, GMSNet \cite{song2020grouped} is trained on both real-world and synthetic noisy/clean pairs to obtaining better generalization performance.

\begin{figure}[!t]
  \vspace{-0ex}
  \begin{center}
    \vspace{-0.0ex}
    \includegraphics[width=0.475\textwidth]{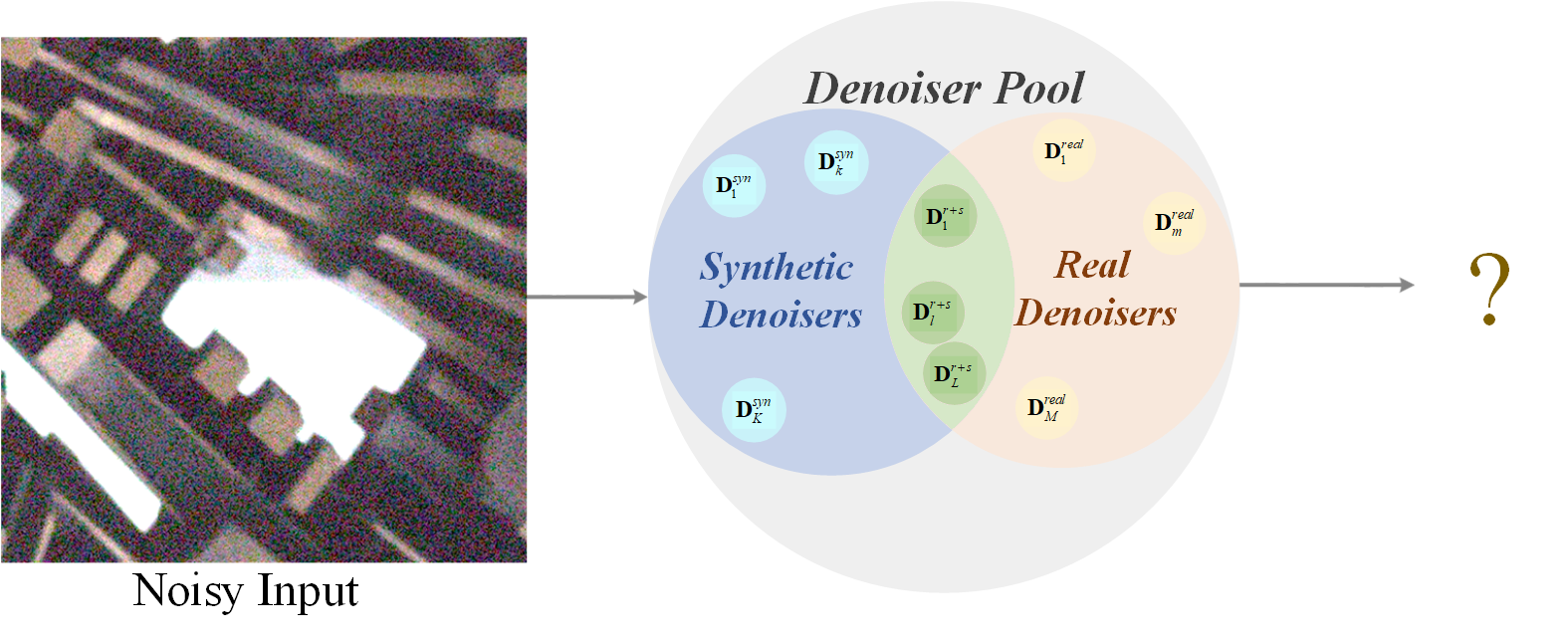}
    \vspace{-0.0ex}
  \end{center}
    \caption{The problem formulation: Given well-trained denoisers with different training settings, how to obtain a robust and better denoised result?}
    \vspace{-0ex}
    \label{fig:problem}
\end{figure}

\begin{figure*}[!htbp]
  \vspace{-0ex}
  \centering
    \subfigure[The pipeline of our proposed ensemble method.]{
      \begin{minipage}[c]{1.\textwidth}
      \centering
        \includegraphics[width=0.99\linewidth]{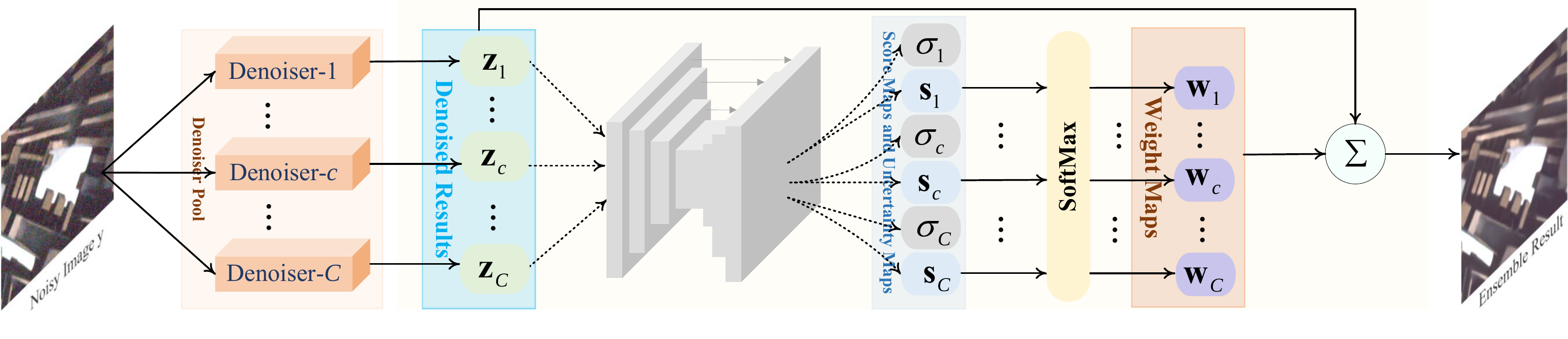}
        \label{fig:wholeall}
      \end{minipage}%
      }
    \vspace{-0.12ex}
    \subfigure[The architecture of our uncertainty scoring network $\mathcal{F}_\Theta$.]{
      \begin{minipage}[c]{1.\textwidth}
      \centering
        \includegraphics[width=0.99\linewidth]{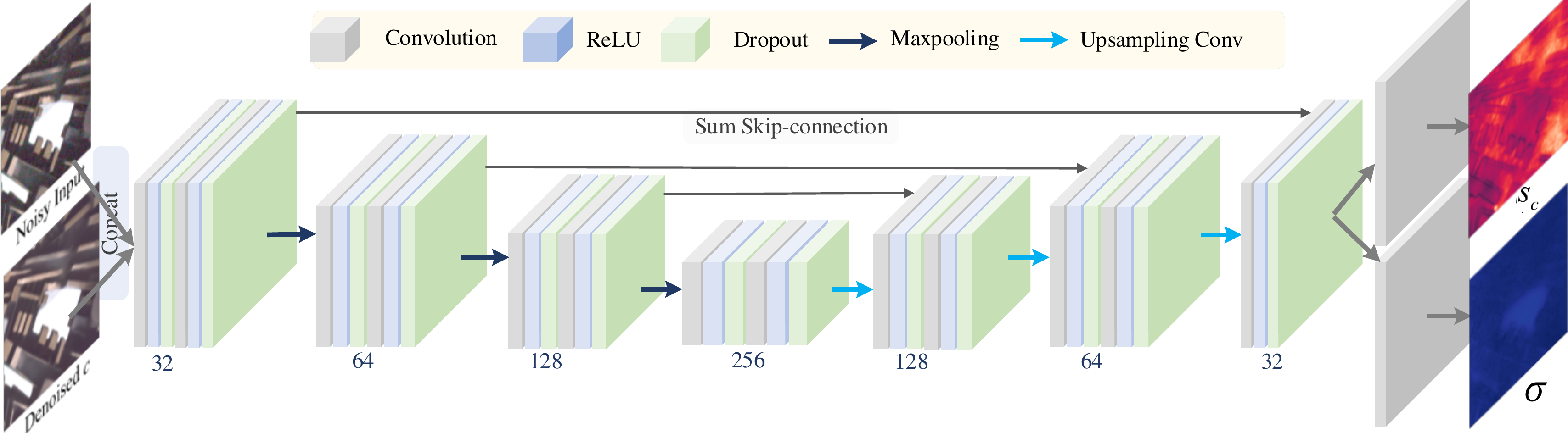}
        \label{fig:uncer}
      \end{minipage}%
      }
    \vspace{-0ex}
    \caption{Overview of our ensemble framework. 
    (a) The overall pipeline of BDE: a noisy image is first processed by $C$ denoisers to obtain denoised images $\mathbf{z}_c$, then the scoring network $\mathcal{F}_\Theta$ takes the noisy image and one of the denoised images as input to estimate the corresponding scoring map $\mathbf{s}_c$ as well as its uncertainty map ${\sigma}_c$, and the final weighting maps $\mathbf{w}_c$ are computed after running all the denoisers' scoring maps by adopting the SoftMax function. 
    (b) The detailed architecture of the uncertainty scoring network. It is a U-Net-based network and adds a dropout layer after ReLU to build MC dropout~\cite{gal2016dropout}. The number of convolution maps in the first level is set to 32, which is doubled after 2$\times$ downsampling. Each level contains two convolution layers in both encoder and decoder to form the 14-layer network.
    }
    \vspace{-0.12ex}
    \label{fig:fusingmeth}
\end{figure*}


Currently, there are several representative deep denoisers have been learned, forming a denoiser pool $\mathbb{D}=\{\mathcal{D}_c\}_{c=1}^C$, as shown in Fig.~\ref{fig:problem}. 
Considering the heterogeneity distribution of real-world noisy images, one denoiser from $\mathbb{D}$ may perform well for some regions while failing for other regions. 
Therefore, we in this work suggest that heterogeneous noises in a real-world noisy image can be separately handled by different denoisers. 
To this end, the denoised result $\hat{\mathbf{z}}_c= \mathcal{D}_c(\mathbf{y})$ can be fused by 
%
\begin{eqnarray} \label{eq:addres}
\mathbf{\hat{x}}  = \! \sum_{c=1}^{C}  \mathbf{w}_c \odot \hat{\mathbf{z}}_c,  \ s.t. \ \mathbf{w}_1 +  ... + \mathbf{w}_C=\mathbf{1},
\end{eqnarray}
where $\mathbf{w}_c$ is the weighting map, and $\odot$ is the entry-wise product. 
Different from \cite{choi2019optimal}, where the image-level ensemble method is proposed for Gaussian denoising, our proposed ensemble strategy would benefit heterogeneous noise. 
As support, our method by fusing different Gaussian noise levels outperforms state-of-the-art CBDNet when applying to real-world noisy images. 
The remaining issue is how to obtain the weighting map, for which we propose a Bayesian learning method.

\subsection{Bayesian Deep Ensemble Denoising Method}\label{fusingmethod}
As shown in Fig.~\ref{fig:wholeall}, our pipeline for estimating the weighting maps contains two parts:
1) estimating the score for each noise removal result by the scoring network $\mathcal{F}$ with parameters $\Theta$
\begin{equation}
	(\mathbf{s}_c, \boldsymbol{\sigma}_c) = \mathcal{F}_\Theta (\mathbf{y}, \hat{\mathbf{z}}_c),
\end{equation}
and 2) obtaining the final weighting maps by the SoftMax function:
\begin{equation} \label{softmax}
(\mathbf{w}_c)_{ij} =  \frac{\exp (\mathbf{s}_c)_{ij}} { \sum^C_{c=1} \exp (\mathbf{s}_c)_{ij}},
\end{equation}
where $\left(i,j\right)$ is the position in the scoring map, and $C$ denotes the number of fused denoisers.
For the scoring network, it takes one of the denoised results $\hat{\mathbf{z}}_c$ with noisy image $\mathbf{y}$ as input and outputs two maps: the scoring map and uncertainty map $\boldsymbol{\sigma}$.
It should be pointed out that the scoring network is based on Bayesian deep learning by considering both input-dependent aleatoric uncertainty and epistemic uncertainty~\cite{kendall2017uncertainties}.
For input-dependent aleatoric uncertainty, it comes from two folds:
i) Both different networks' architectures and different training sets lead to various denoised results.
ii) The unknown test set with unknown-type noise leads to completely unknown effects.
Epistemic uncertainty is modeled by placing a prior distribution (\emph{e.g.} Gaussian prior distribution) over a model’s weights, and then trying to capture how much these weights vary given some data.
Here, we assume a Gaussian prior distribution with zero mean over CNN's weights, and estimate the weighting uncertainty $\boldsymbol{\sigma}$ (\emph{i.e.} variance) with the scoring map to capture how much distribution noise we have in the output scoring maps.
Note that $\boldsymbol{\sigma}$ is only utilized to train the robust scoring network and abounded during the testing phase.

\subsubsection{Network Configuration}
%
As shown in Fig~\ref{fig:uncer}, our scoring network is a 4-level U-Net architecture with 3 sum skip connections: 
i) To model input-dependent aleatoric uncertainty, we add the dropout layer after each ReLU layer to build MC dropout~\cite{gal2016dropout}.
ii)  All the convolutional kernel size is set to $3\times3$.
iii) In the first level, the feature map is set to 32, and doubled after max-pooling.
iv) Each level of the network contains two convolution layers in both encoder and  decoder, and 
the final layer of the scoring network is 14.

\subsubsection{Learning Objective}

Once obtaining all the weighting maps, we can get the fused results by $\mathbf{\hat{x}}= \sum_c^C \mathbf{{w}}_c \odot \hat{\mathbf{z}}_c$. 
And then it is natural to impose supervision based on the final fused denoising result,
\begin{eqnarray} \label{eq:sum}
	\mathcal{L}_{fuse} = \underset{\Theta}{\min}   \| \sum_{c=1}^C \mathbf{{w}}_c \odot \hat{\mathbf{z}}_c - \mathbf{x}^{gt} \|_1.
\end{eqnarray}

Besides, we suggest introducing supervision on weighting map. 
The ground-truth weighting map can be generated by two approaches:
i) utilizing the convex optimization in \cite{choi2019optimal};
ii) selecting the best one and then setting its mask to 1 while the others to zeros. 
However, the first approach couldn't generate pixel-wise ground-truth weight in real time.
Hence, we adopt the second approach for generating the ground-truth weights, as follows:
\begin{align*}
\label{maskgt}
(\mathbf{w}_c^{gt})_{ij} =\left\{
\begin{aligned}
1 & , \ \ \text{if} \ \ (\mathbf{e}_c)_{ij}= \min\{ (\mathbf{e}_c)_{ij}\}_1^C, \\
0 & , \ \ \text{otherwise},
\end{aligned}
\right.
\end{align*}
where $(\mathbf{e}_c)_{ij}$ is the $i,j$-th pixel MSE between $c$-th prediction result and ground-truth, \emph{i.e.} $(\mathbf{e}_c)_{ij}= \| \left(\hat{\mathbf{z}}_c\right)_{ij} - \left(\mathbf{x}^{gt}\right)_{ij} \|^2$.
Referring to \cite{kendall2017uncertainties}, we adopt the following negative log-likelihood loss  
\begin{equation} \label{eq:nlltv}
\begin{aligned}
\mathcal{L}_{NLL} = \underset{\Theta}{\min} &\frac{1}{C} \sum_{c=1}^C  \frac{1}{{ \boldsymbol{\sigma}}^{2}_c} \| \mathbf{{w}}_c - \mathbf{w}_c^{gt} \|_2 \\&+ \log \left({\boldsymbol{\sigma}}_c \right)  + \lambda_{TV} TV\left( \mathbf{w}_c \right),
\end{aligned}
\end{equation}
where $\hat{\boldsymbol{\sigma}}$ is uncertainty map, $TV\left( \cdot \right)$ denotes the Total Variation regularization to smooth the estimated weighting map and $\lambda_{TV}$ is the trade-off parameter.

Finally, we can get the total loss function for learning $\mathcal{F}_\Theta$
\begin{eqnarray} \label{eq:weights}
\mathcal{L} = \mathcal{L}_{NLL} + \lambda_0 \mathcal{L}_{fuse} ,
\end{eqnarray}
where $\lambda_0$ controls the weighting between the two loss terms.

\begin{figure*}[!htbp]
     
      \begin{minipage}[c]{0.137\textwidth}
      \vspace{0.5ex}
      \centering
        \centerline{\includegraphics[width=0.99\linewidth]{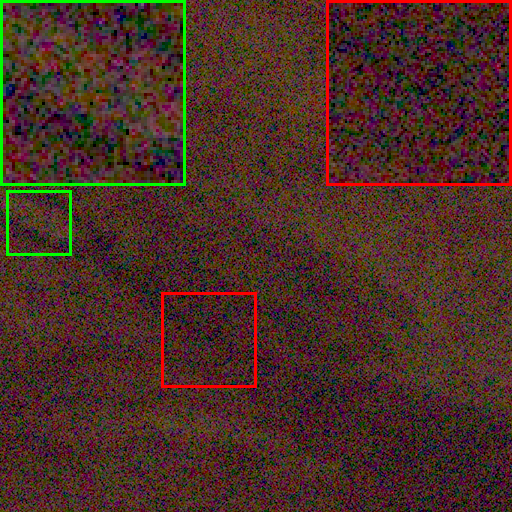}}
        \centerline{Noisy Image}
        \label{fig:noisya2}
      \end{minipage}%
    \hspace{0.ex}
      \begin{minipage}[c]{0.137\textwidth}
      \centering
        \centerline{\includegraphics[width=0.99\linewidth]{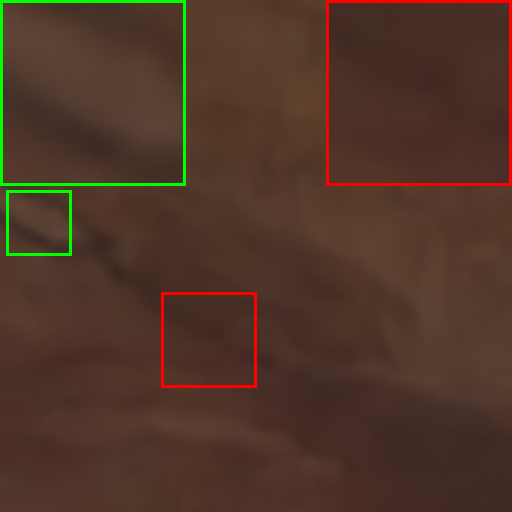}}
        \centerline{CBDNet~\cite{guo2019toward}}
        \label{fig:CBDNet}
      \end{minipage}%
    \hspace{-0.ex}
      \begin{minipage}[c]{0.137\textwidth}
      \centering
        \centerline{\includegraphics[width=0.99\linewidth]{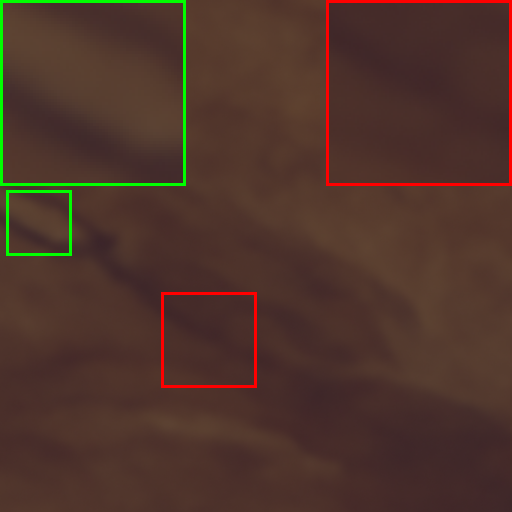}}
        \centerline{NBNet~\cite{cheng2021nbnet}}
        \label{fig:NBNet}
      \end{minipage}%
    \hspace{-0.ex}
      \begin{minipage}[c]{0.137\textwidth}
      \centering
        \centerline{\includegraphics[width=0.99\linewidth]{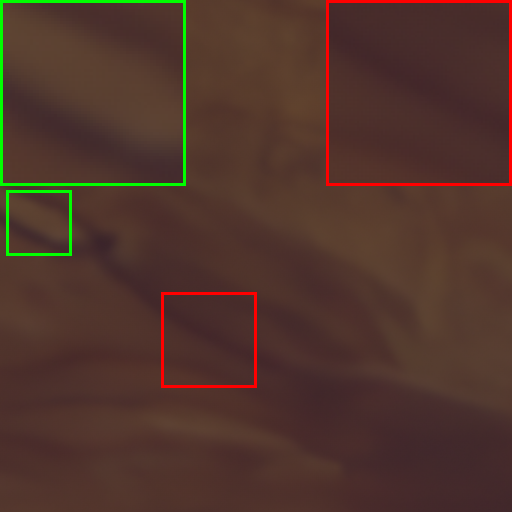}}
        \centerline{HINet~\cite{chen2021hinet}}
        \label{fig:HINet}
      \end{minipage}%
    \hspace{-0.ex}
      \begin{minipage}[c]{0.137\textwidth}
      \centering
        \centerline{\includegraphics[width=0.99\linewidth]{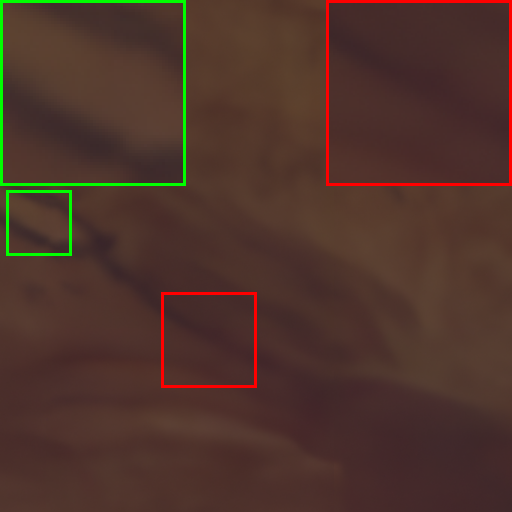}}
        \centerline{Uformer~\cite{wang2021uformer}}
        \label{fig:Uformer}
      \end{minipage}%
    \hspace{-0.ex}
      \begin{minipage}[c]{0.137\textwidth}
      \centering
        \centerline{\includegraphics[width=0.99\linewidth]{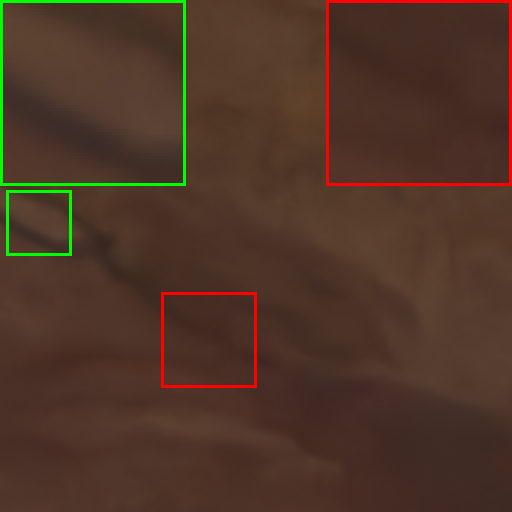}}
        \centerline{GMSNet~\cite{song2020grouped}}
        \label{fig:GMSNet}
      \end{minipage}%
    \hspace{-0.ex}
      \begin{minipage}[c]{0.137\textwidth}
      \centering
        \centerline{\includegraphics[width=0.99\linewidth]{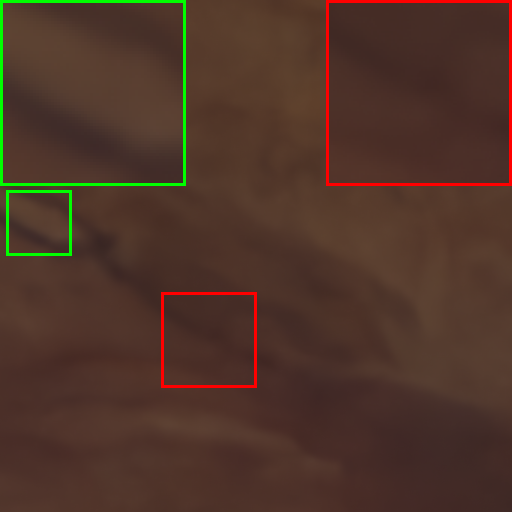}}
        \centerline{BDE}
        \label{fig:23}
      \end{minipage}%
    \vspace{0ex}
      \begin{minipage}[c]{0.137\textwidth}
      \centering
        \centerline{\includegraphics[width=0.99\linewidth]{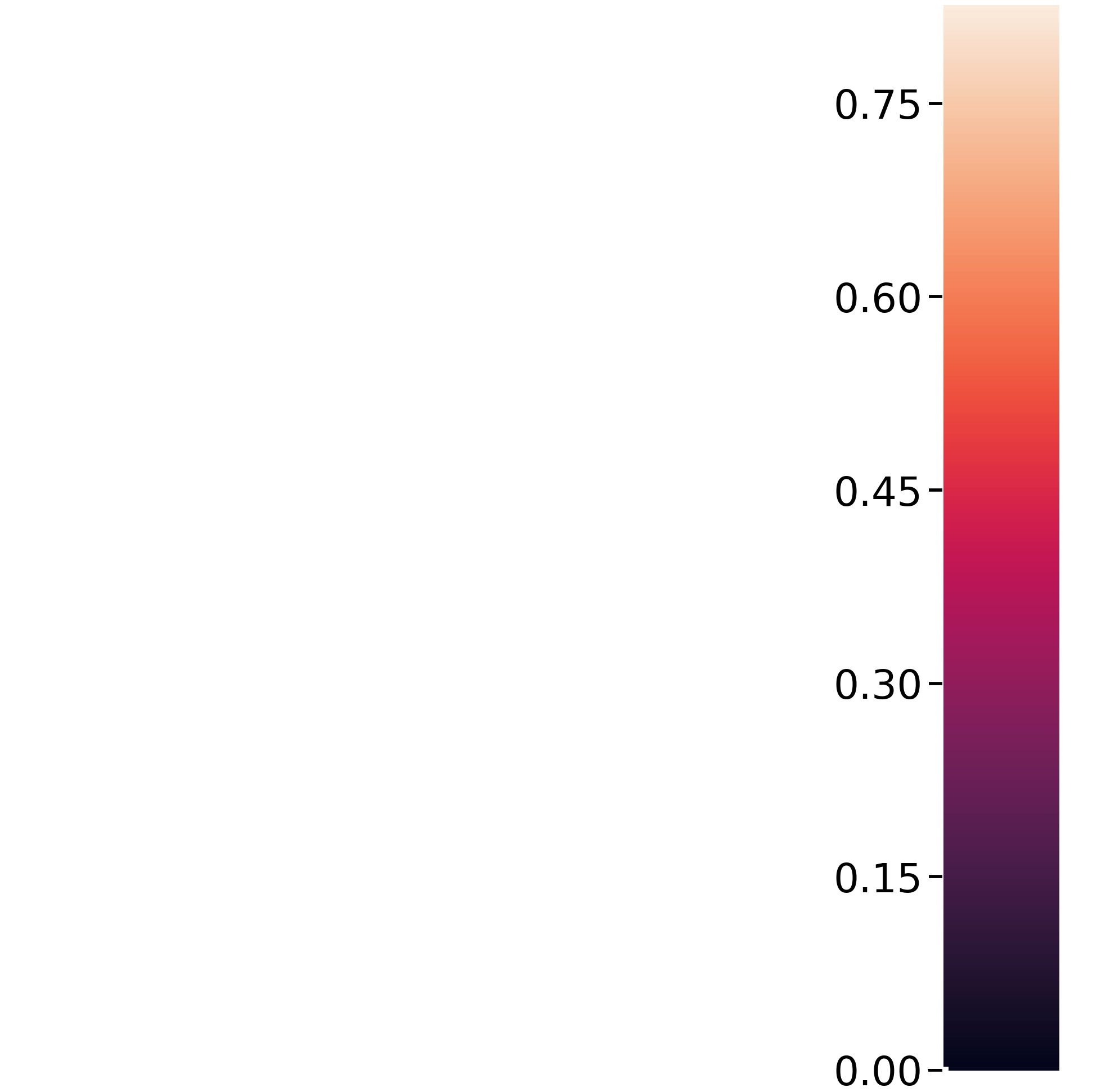}}
        \label{fig:23_43v}
      \end{minipage}%
    \hspace{-0.ex}
      \begin{minipage}[c]{0.137\textwidth}
      \centering
        \centerline{\includegraphics[width=0.99\linewidth]{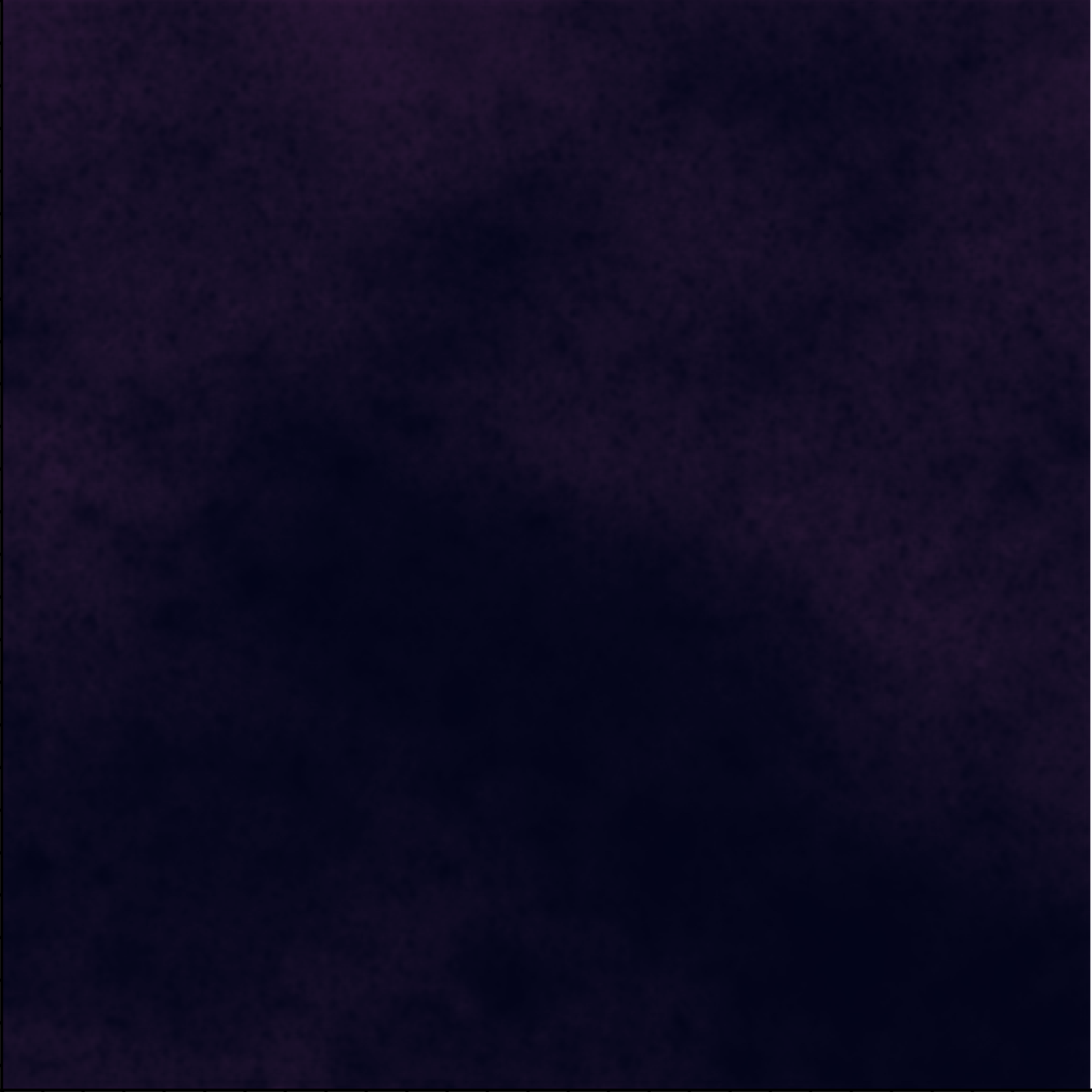}}
        \label{fig:CBDNet34v}
      \end{minipage}%
    \hspace{-0.ex}
      \begin{minipage}[c]{0.137\textwidth}
      \centering
        \centerline{\includegraphics[width=0.99\linewidth]{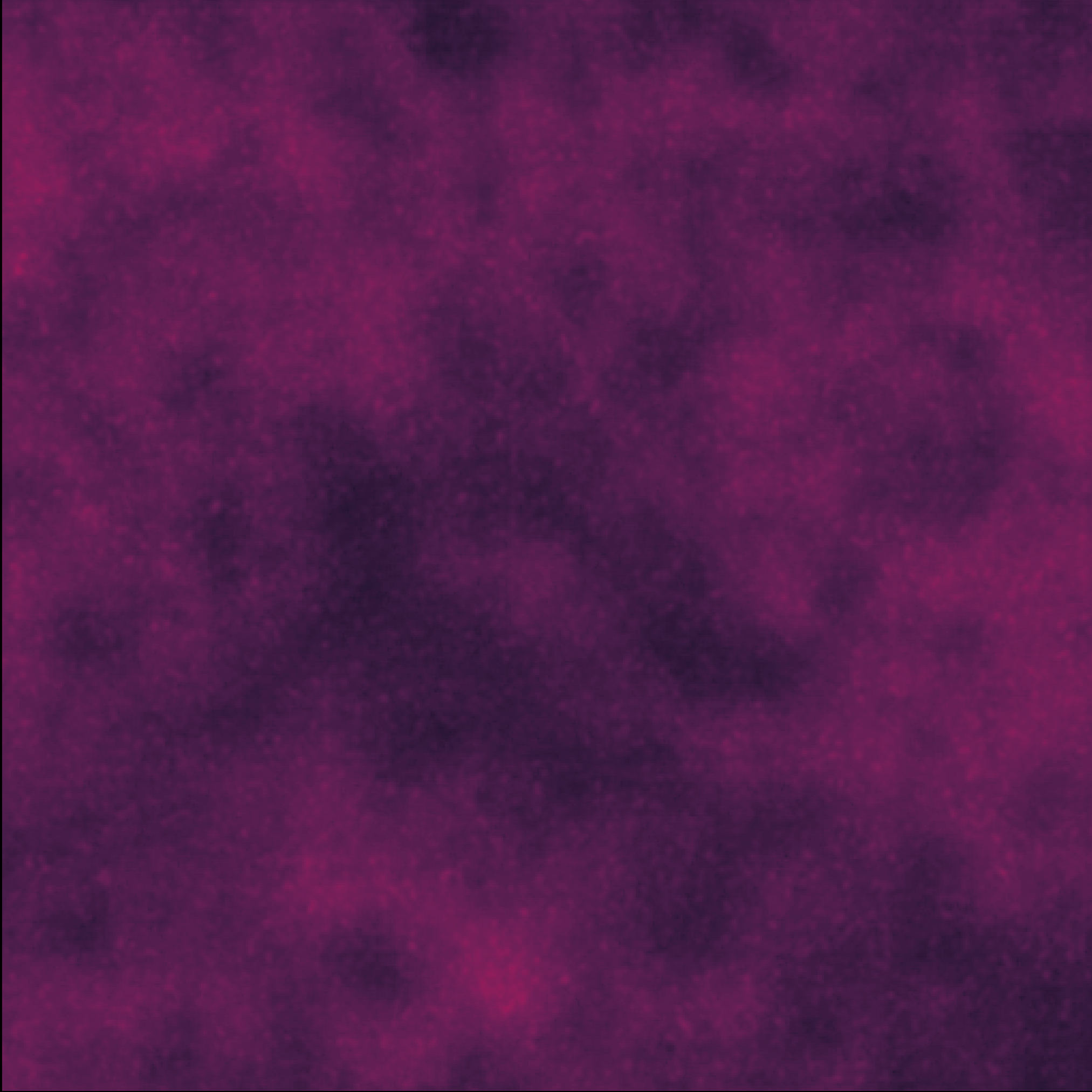}}
        \label{fig:NBNet34v}
      \end{minipage}%
    \hspace{-0.ex}
      \begin{minipage}[c]{0.137\textwidth}
      \centering
        \centerline{\includegraphics[width=0.99\linewidth]{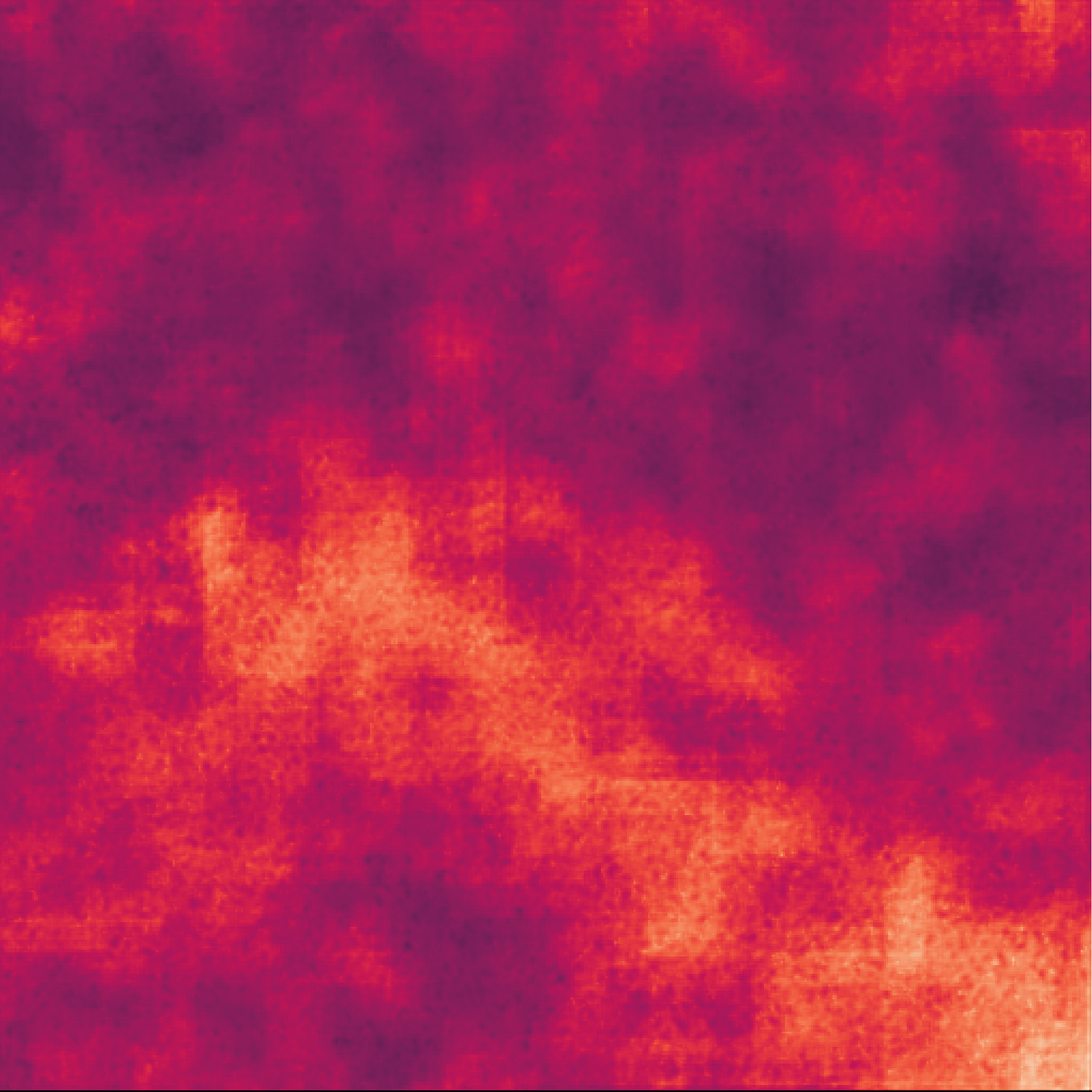}}
        \label{fig:HINet34v}
      \end{minipage}%
    \hspace{-0.ex}
      \begin{minipage}[c]{0.137\textwidth}
      \centering
        \centerline{\includegraphics[width=0.99\linewidth]{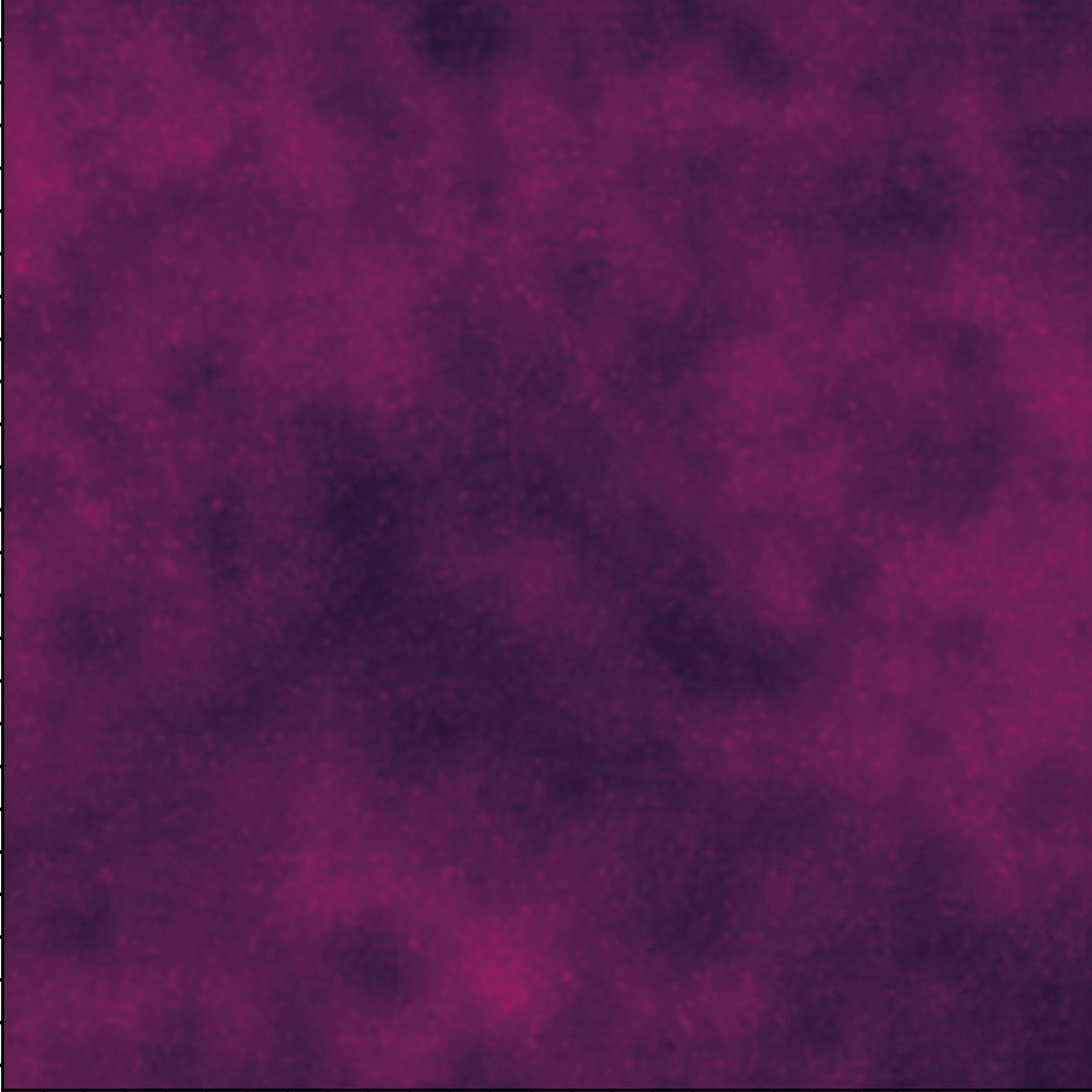}}
        \label{fig:Uformer34v}
      \end{minipage}%
     \hspace{-0.ex}
        \begin{minipage}[c]{0.137\textwidth}
        \centering
          \centerline{\includegraphics[width=0.99\linewidth]{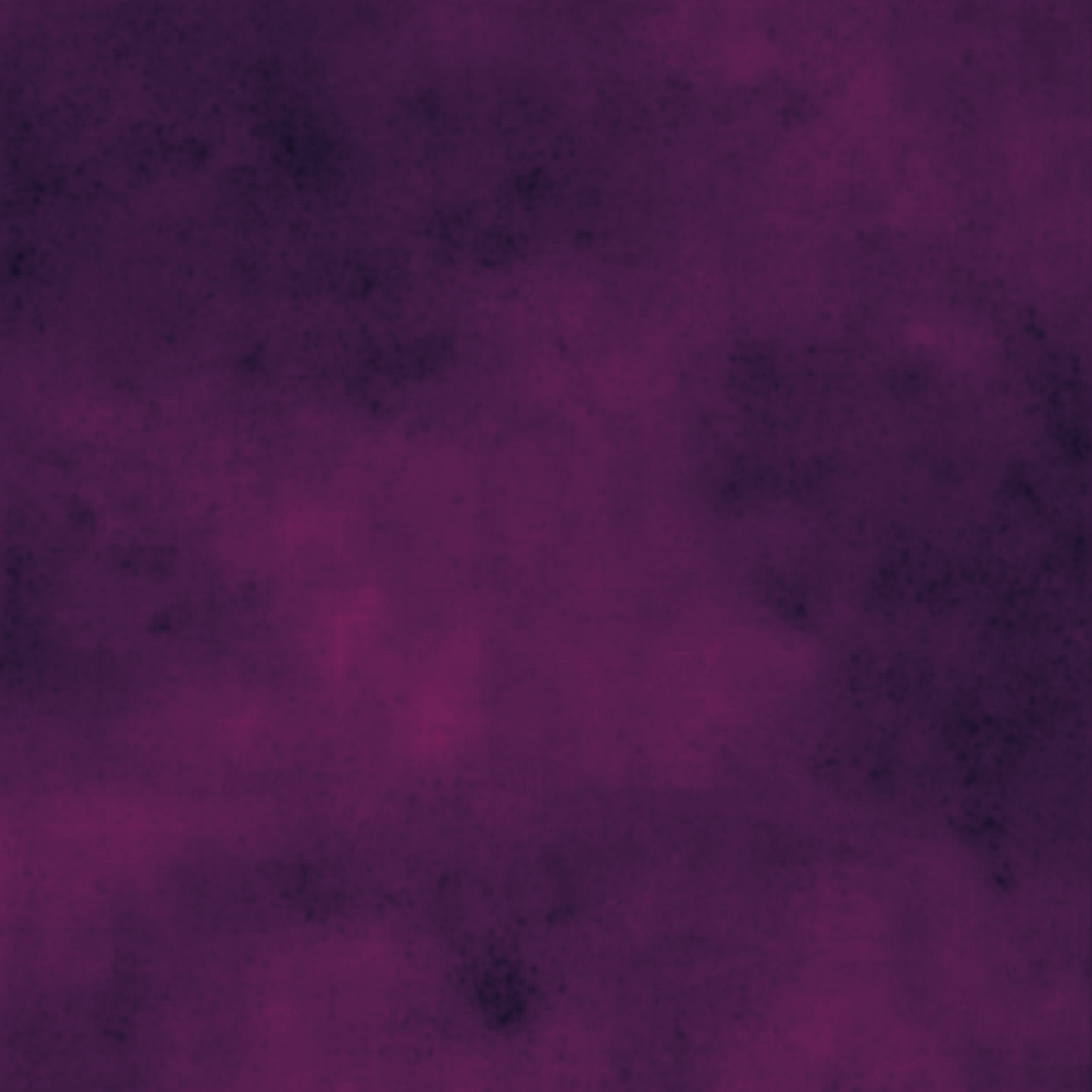}}
          \label{fig:GMSNet34v}
        \end{minipage}%
      \hspace{-0.ex}
      \begin{minipage}[c]{0.137\textwidth}
      \centering
        \centerline{\includegraphics[width=0.99\linewidth]{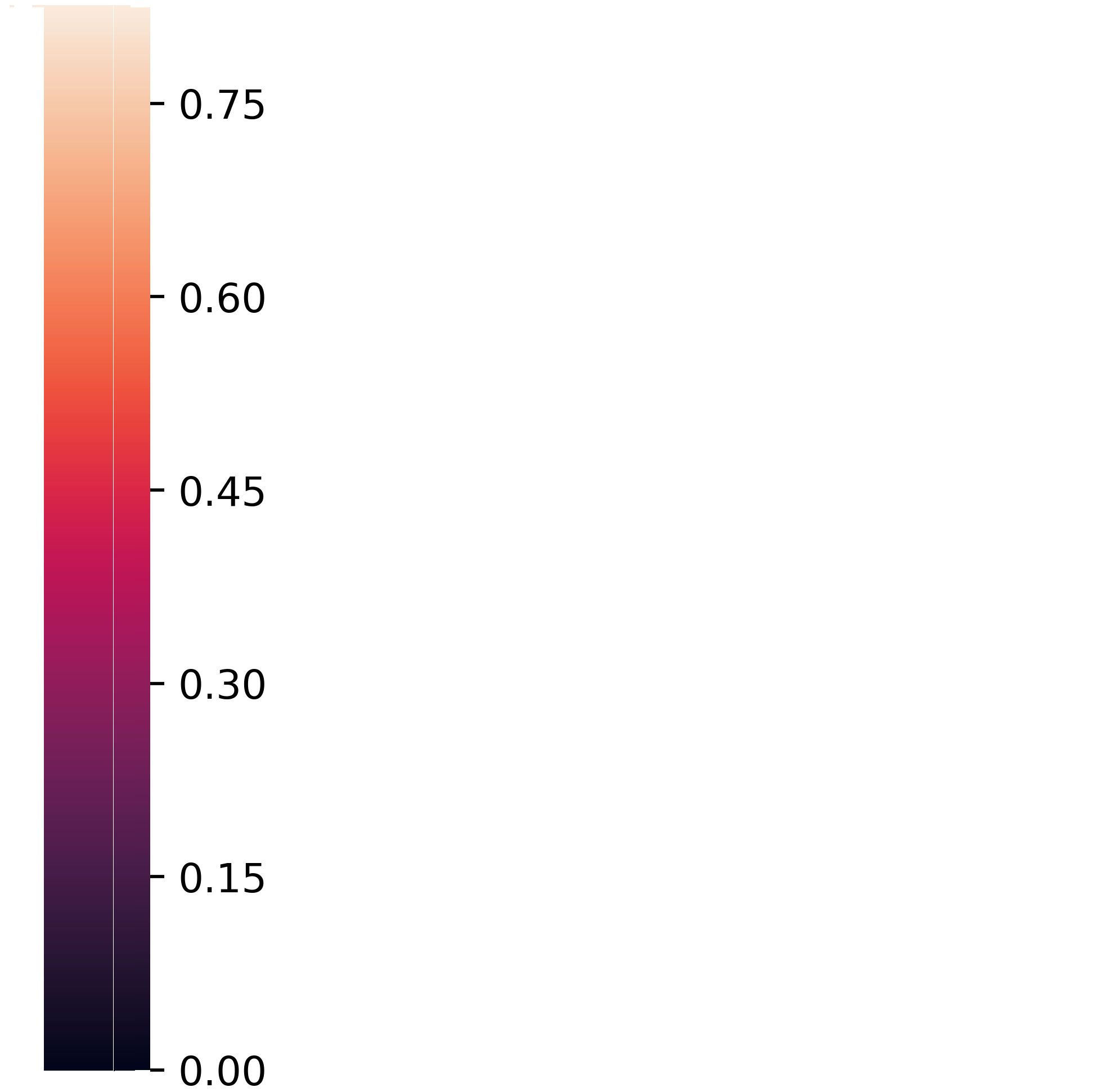}}
        \label{fig:2343v}
      \end{minipage}%
        \vspace{0ex}
      \begin{minipage}[c]{0.137\textwidth}
      \centering
        \centerline{\includegraphics[width=0.99\linewidth]{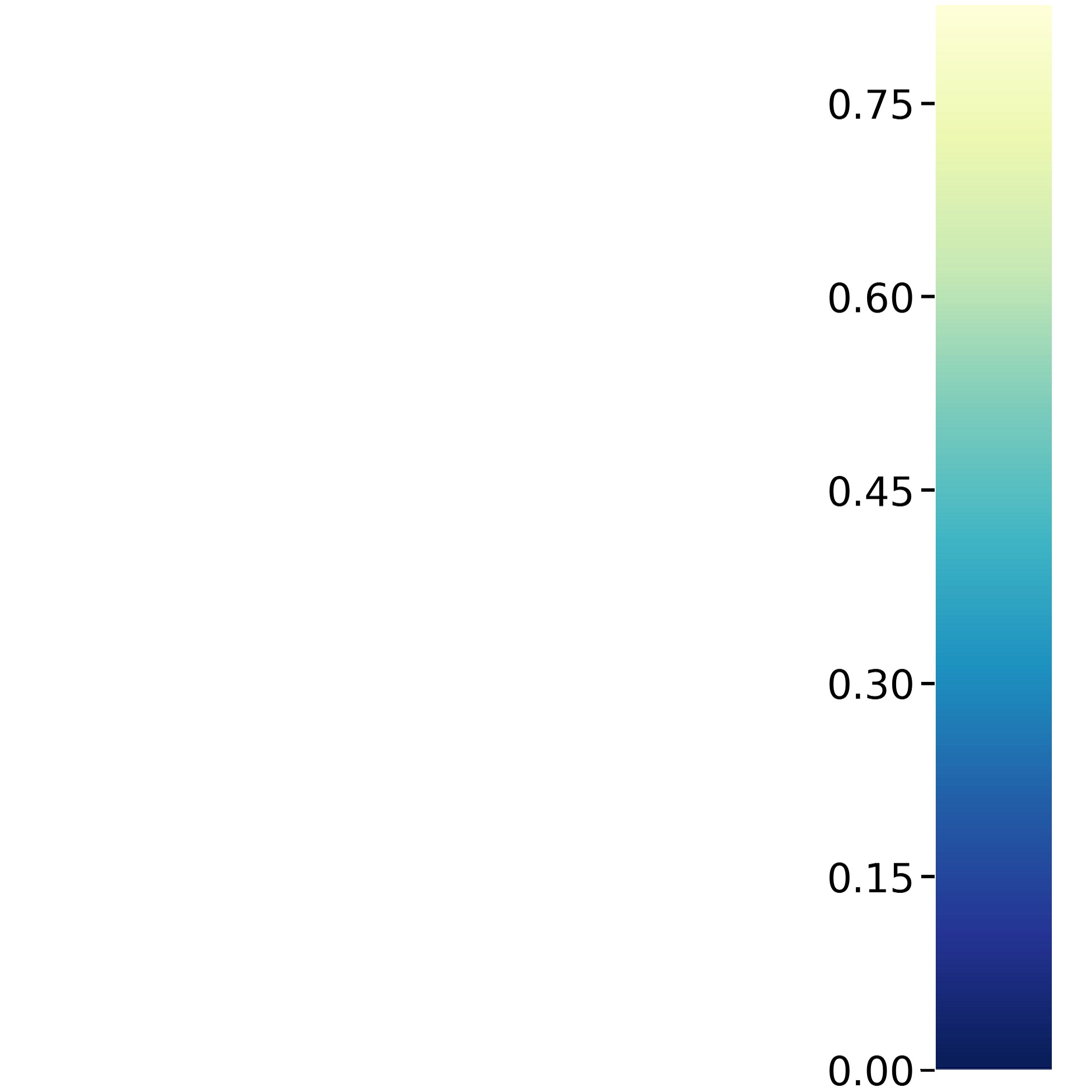}}
        \label{fig:23_43u}
      \end{minipage}%
    \hspace{-0.ex}
      \begin{minipage}[c]{0.137\textwidth}
      \centering
        \centerline{\includegraphics[width=0.99\linewidth]{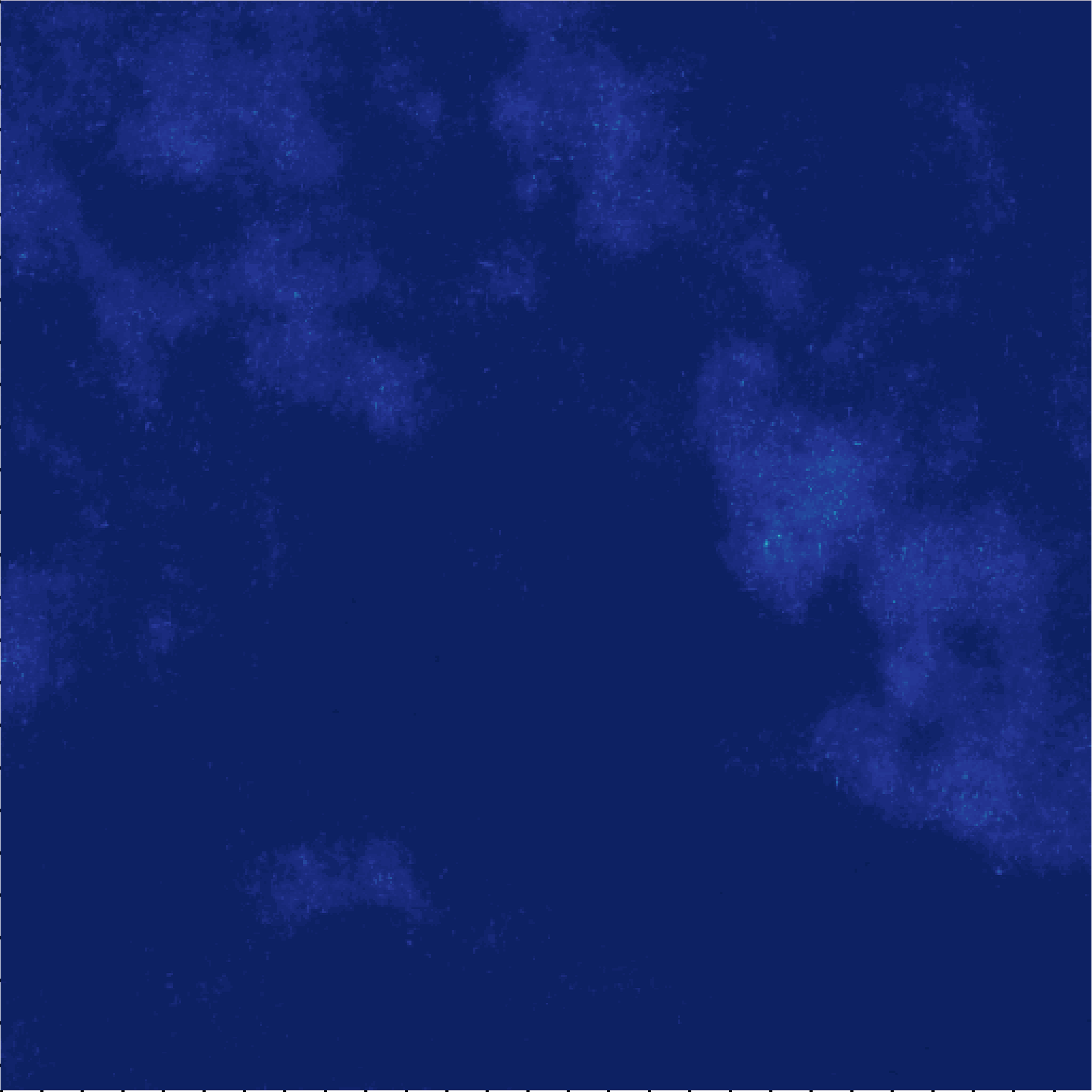}}
        \label{fig:CBDNet34u}
      \end{minipage}%
    \hspace{-0.ex}
      \begin{minipage}[c]{0.137\textwidth}
      \centering
        \centerline{\includegraphics[width=0.99\linewidth]{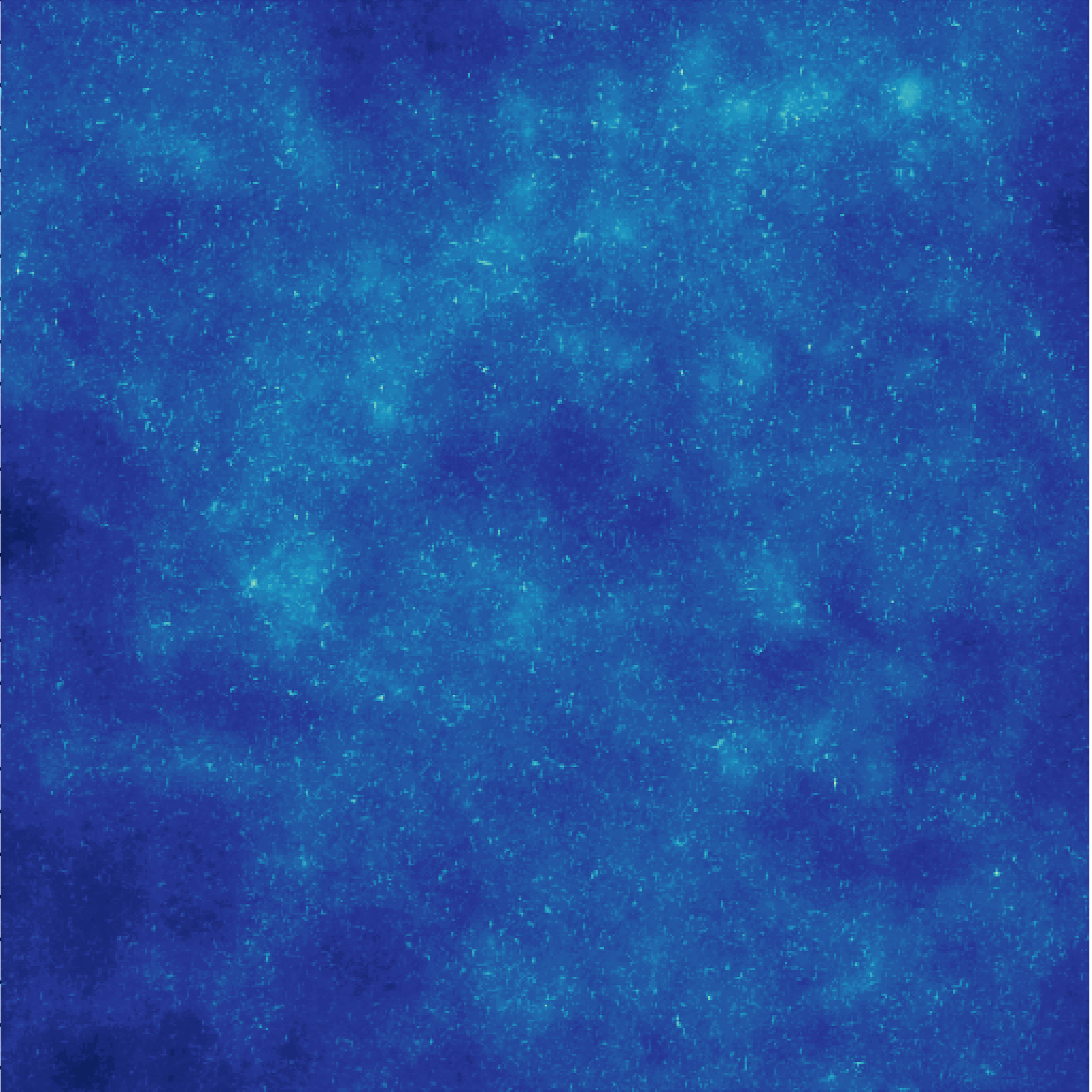}}
        \label{fig:NBNet34u}
      \end{minipage}%
    \hspace{-0.ex}
      \begin{minipage}[c]{0.137\textwidth}
      \centering
        \centerline{\includegraphics[width=0.99\linewidth]{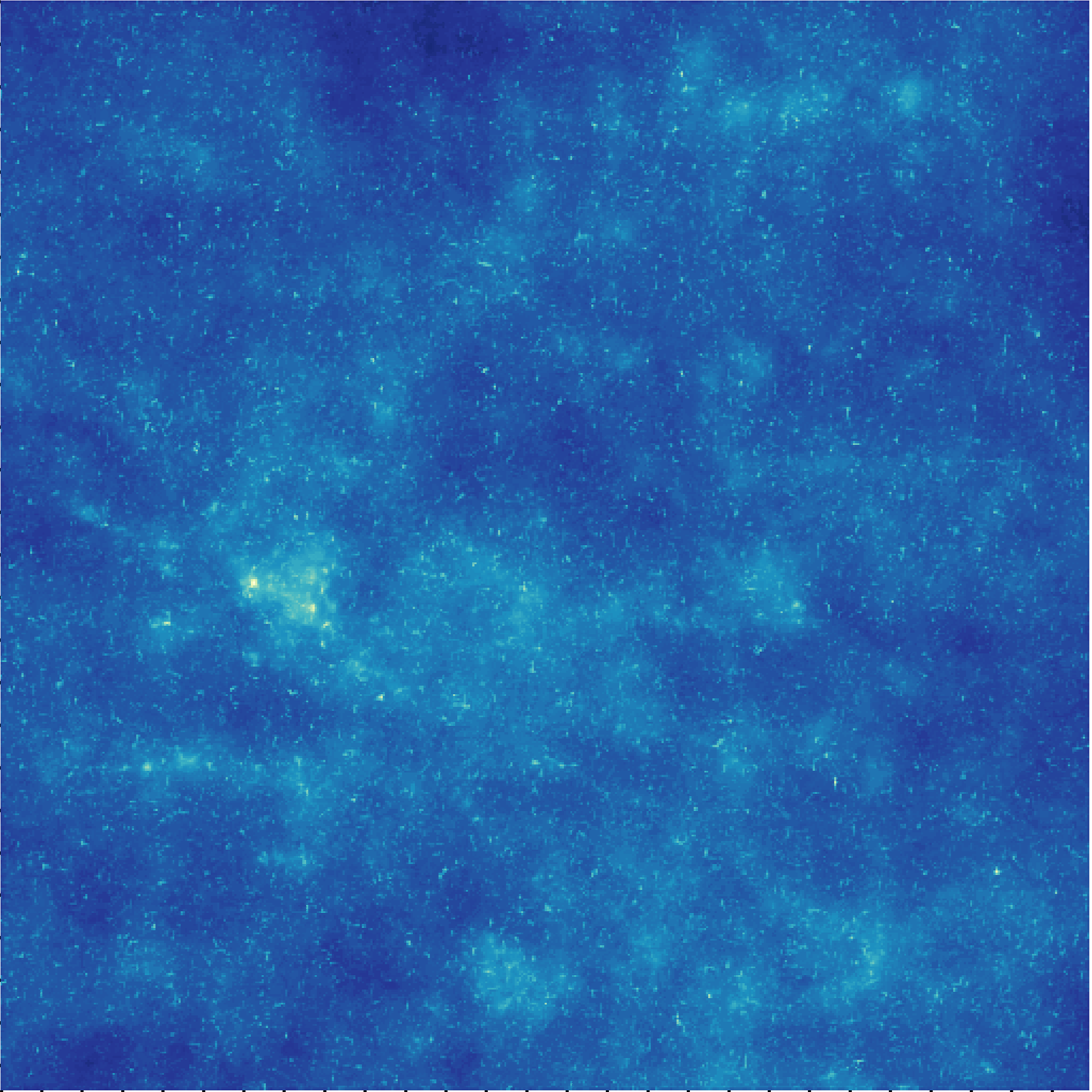}}
        \label{fig:HINet34u}
      \end{minipage}%
    \hspace{-0.ex}
      \begin{minipage}[c]{0.137\textwidth}
      \centering
        \centerline{\includegraphics[width=0.99\linewidth]{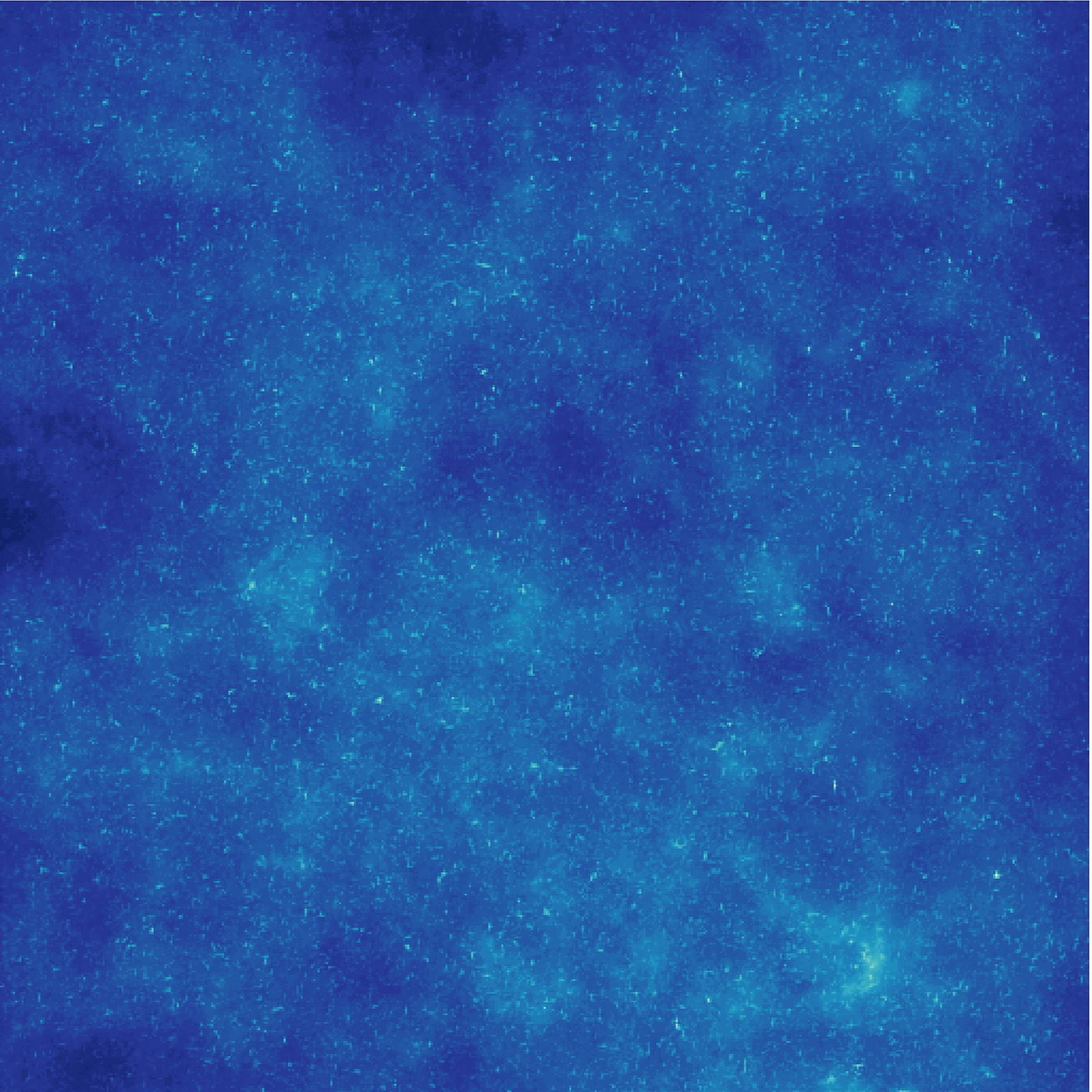}}
        \label{fig:Uformer34u}
      \end{minipage}%
     \hspace{-0.ex}
        \begin{minipage}[c]{0.137\textwidth}
        \centering
          \centerline{\includegraphics[width=0.99\linewidth]{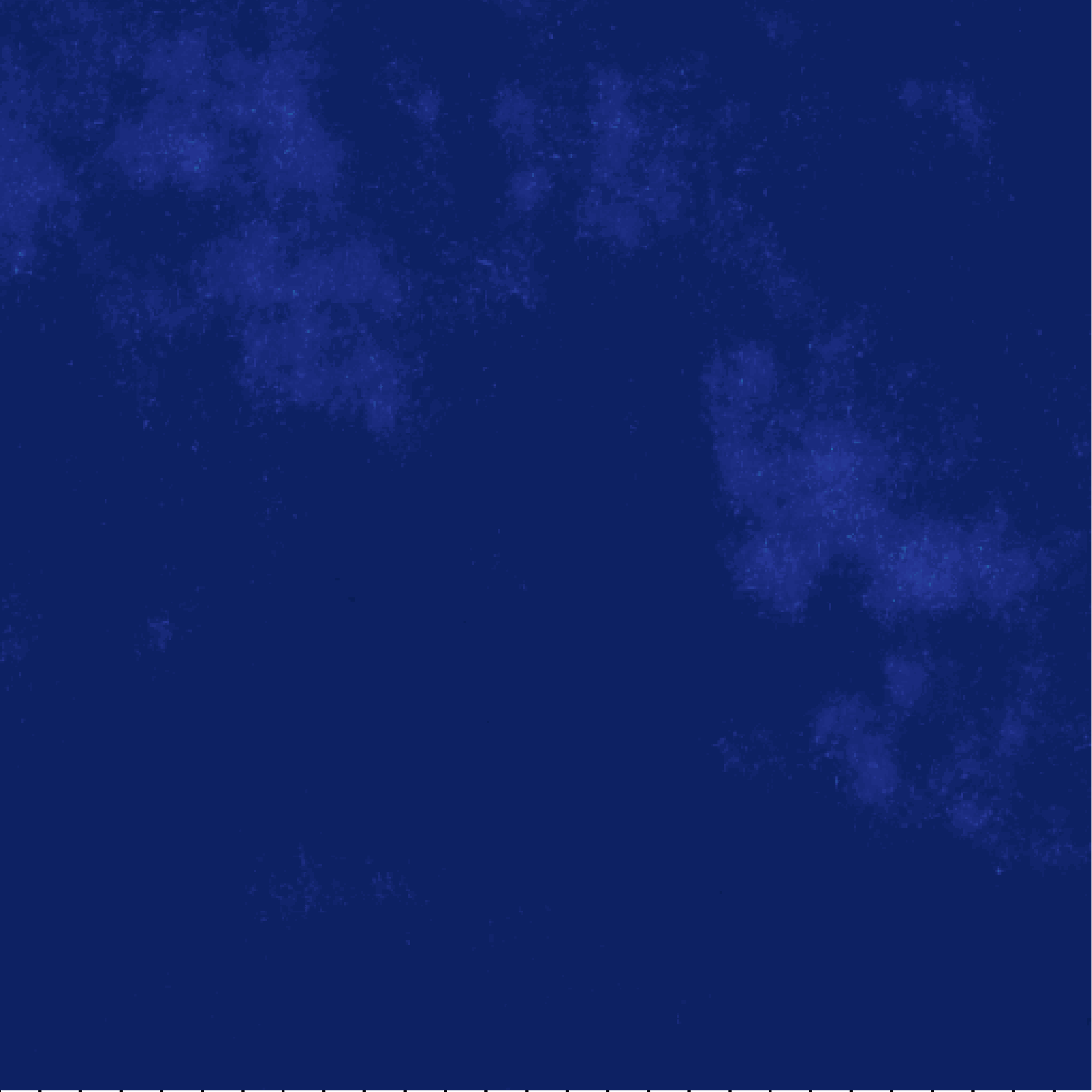}}
          \label{fig:GMSNet34u}
        \end{minipage}%
      \hspace{-0.ex}
      \begin{minipage}[c]{0.137\textwidth}
      \centering
        \centerline{\includegraphics[width=0.99\linewidth]{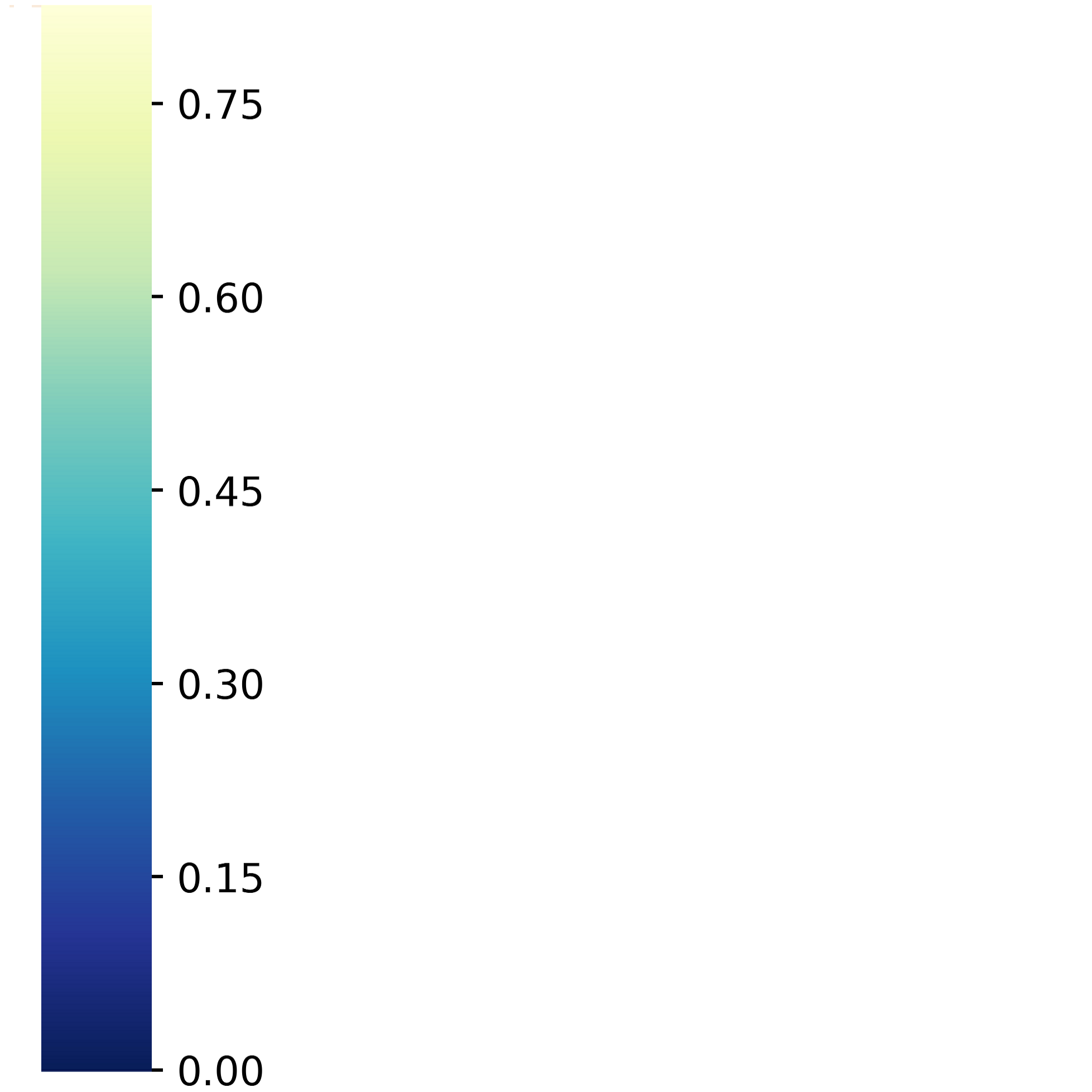}}
        \label{fig:2343u}
      \end{minipage}%
    \vspace{-0.ex}
      \caption{ The visual results of different denoisers on DND benchmark: denoising results by five state-of-the-art denoisers are in the first row, weighting maps are in the second row, and uncertainty maps are in the third row.
      It can be easily found that our proposed method is region-aware to fuse different noise removal results. 
      Our result in the red box is selected from GMSNet, while the result in the green box is obtained by linearly combining the results of five denoisers.
      }
      \label{fig:dndv}
      \vspace{-0.2ex}
    \end{figure*}
    
\begin{table*}[!t]\footnotesize
	\centering
	\caption{
	  {
	Quantitative results of PSNR(dB)/SSIM on DND benchmark.
	BDE takes five denoisers, \emph{i.e.}  CBDNet~\cite{guo2019toward}, NBNet~\cite{cheng2021nbnet},  GMSNet~\cite{song2020grouped},  Uformer~\cite{wang2021uformer} and HINet~\cite{chen2021hinet}, as denoiser pool.
	ModelH is a big model that combines the fused denoisers and increases convolution channels to keep the similar Flops and parameters of the overall BDE model.
	{\bf{Bold}} font indicates the best performance.}
	}
	\vspace{-0.0in}
	\renewcommand\tabcolsep{1.0pt}
	\renewcommand\arraystretch{1.2}
	\scalebox{0.8051}{
	\begin{tabular}{|p{1.3cm}<{\centering} || p{0.85cm}<{\centering} | p{1.6cm}<{\centering} | p{1.68cm}<{\centering} | p{1.68cm}<{\centering} | p{2.008cm}<{\centering} | p{1.68cm}<{\centering} | p{1.68cm}<{\centering} | p{1.48cm}<{\centering} |  p{1.48cm}<{\centering} | p{1.498cm}<{\centering} | p{1.70cm}<{\centering} | p{1.28cm}<{\centering} | p{1.28cm}<{\centering} | p{1.58cm}<{\centering} |  p{1.3cm}<{\centering} | p{1.3cm}<{\centering} |}
	\hlineB{2.2}
	\rowcolor{lightgray}
	Dataset& Metric & BM3D~\cite{dabov2007image} & CBDNet~\cite{guo2019toward}& CycleISP~\cite{zamir2020cycleisp} & Path-restore~\cite{yu2021path} & MPRNet~\cite{mehri2021mprnet} & MIRNet~\cite{zamir2020learning} & NBNet~\cite{cheng2021nbnet} & HINet~\cite{chen2021hinet} &  Uformer~\cite{wang2021uformer} &  
	GMSNet~\cite{song2020grouped} & ModelH & BDE  \\ \hline \hline
		\multirow{2}{*}{DND~\cite{plotz2017benchmarking}} 
	& PSNR  & 34.51 & 38.06 & 39.56  & 39.72 & 39.82 & 39.88 & 39.89 & 39.90 & 39.96  & 40.24  & 40.36 & \bf{40.52}   \\ 
	& SSIM  & 0.951 & 0.942 & 0.956 & 0.959 & 0.954 & 0.956 & 0.955 & 0.955 &  0.956  & 0.962 & 0.962  & \bf{0.964}     \\
	  \hlineB{2.2}
	\end{tabular}
	}
	\vspace{-0.12in}
	\label{tab:dnd}
  \end{table*}

\subsection{Discussion}
Here, we mainly focus on analyzing our first stage, which merges different denoisers' results to obtain preliminary noise removal results. 
From different views, it can help to more easily understand our proposed method.

\subsubsection{Relationship with path-restore}
In  \cite{yu2021path}, Yu \emph{et al.} restore the corrupted image by a path-restore CNN method that can dynamically select an appropriate route for each image region.
The pathfinder in this path-restore CNN method can provide the thumbnail masks for selecting different features from the previous layer.
As shown in Fig.~\ref{fig:fusingmeth}, our proposed ensemble method predicts the weighting maps to select or combine different regions of denoised results.
In this view, our proposed can be seen as a region-aware method that is similar to the path-restore CNN method.
Different from \cite{yu2021path}, our proposed method can merge infinite denoisers while the paths in the path-restore CNN method are fixed.
Moreover, our predicted probability map is a pixel-wise map instead of the thumbnail one, which can provide more detailed region-aware information.

 \subsubsection{Relationship with dictionary learning}
 In dictionary learning, the restored image signal $\mathbf{\hat{x}}$ can be obtained a linear combination of atomic bases, \emph{i.e.} $\mathbf{\hat{x}}=\mathbf{d}_{n} \mathbf{w}$, where $\mathbf{d}_{n}$ represents the dictionary atom set and $\mathbf{w}$ is the corresponding coefficient vector~\cite{zheng2021deep}. 
 If atomic bases are different denoisers, it is very similar to our ensemble method in Eq.(\ref{eq:addres}), which is also a linear combination function.
 However, in our proposed method, the atoms, \emph{i.e.} different denoisers, are well-trained and then fixed, the coefficient $\mathbf{w}$  can be solved by training a CNN and then fast inference in the testing phase. 
 While $\mathbf{d}_{n}$ and $\mathbf{w}$ in dictionary learning are learned by alternating direction method of multipliers (ADMM) algorithm.
 During the test phase, our ensemble method can be very fast by GPU acceleration, but the dictionary learning is often extremely time-consuming. 
 Moreover, our BDE is hopeful to fuse infinite denoisers rather than the fixed number.
 Above all, our BDE can be seen as an improved dictionary learning method.
  

 \subsection{Extension to Other Image Restoration Tasks}
 Although our BDE is proposed for image denoising, it is potential to be beneficial for other image restoration tasks. 
 In this work, we take image deblurring, image deraining and single image super-resolution as examples. 
 By selecting several well-trained corresponding restoration models into pool $\mathbb{D}$, the learning and inference of BDE for these tasks are exactly same with image denoising. 
 Although degradation in these tasks may be spatially dependent, pixel-wise ensemble by our BDE is also effective, and can notably improve the performance on benchmark datasets (Referring to Sec. \ref{sec:experiments extension}).

\section{Experiments}~\label{secIV}
In this section, we first provide the detailed implementation of our proposed method, such as training sets and testing sets.
Then different results and the ablation study of our method are displayed for demonstrating the effectiveness of our proposed BDE method.
For all the compared methods, we download their source codes and well-trained models from the corresponding authors’ websites. 
PSNR and SSIM  are utilized as basic metrics for evaluating the performance of different methods. 
And the visual results are also provided for quantitative comparison. 

\begin{table*}[!htbp]\footnotesize
\centering
\caption{
The parameters,  Flops, and running times of different compared methods.
Note that the Flops are tested on $256 \times 256$ patch.
ModelH combines the fused denoisrs, \emph{i.e.}  CBDNet~\cite{guo2019toward}, NBNet~\cite{cheng2021nbnet},  GMSNet~\cite{song2020grouped},  Uformer~\cite{wang2021uformer}, and HINet~\cite{chen2021hinet}, and increases convolution channels to match the same Flops and parameters of the overall model.
}
\vspace{-0.02in}
\renewcommand\tabcolsep{1.0pt}
\renewcommand\arraystretch{1.0}
\scalebox{1.0}{
\begin{tabular}{|p{3.19cm}<{\centering}|| p{1.98cm}<{\centering}| p{1.98cm}<{\centering}| p{1.98cm}<{\centering} | p{1.98cm}<{\centering} | p{1.98cm}<{\centering} | p{1.98cm}<{\centering}| p{1.98cm}<{\centering}|}
\hlineB{2.5}
\rowcolor{lightgray} Performance & CBDNet~\cite{guo2019toward} & NBNet~\cite{cheng2021nbnet} & Uformer~\cite{wang2021uformer} & HINet~\cite{chen2021hinet} & GMSNet~\cite{song2020grouped} & ModelH & BDE \\ \hline  \hline
$\#$Param. ($\times10^6$, M)    & 4.365   &  10.45  &   20.58 & 88.65 &  5.074 & 140.27 & 136.37   \\
\rowcolor{tinygray} Flops ($\times10^9$, G)    & 40.33    &   257.93 &   44.46 & 170.71 &  331.91 & 920.39 & 893.34   \\
Running Time  (s)  &   0.009 &  0.045  & 0.068   & 0.021 &  0.106 & 0.283 &  0.259  \\
\hlineB{1.5}
\end{tabular}
}
\label{tab:factor}
\vspace{-0.05in}
\end{table*}
\begin{table*}[!htbp]\footnotesize
  \centering
  \caption{
    {
  Quantitative results of PSNR(dB)/SSIM on SIDD validation benchmark.
  BDE takes five denoisers, \emph{i.e.}  CBDNet~\cite{guo2019toward}, NBNet~\cite{cheng2021nbnet},  GMSNet~\cite{song2020grouped},  Uformer~\cite{wang2021uformer}, and HINet~\cite{chen2021hinet}, as denoiser pool.
  {\bf{Bold}} font indicates the best performance.}
  }
  \vspace{-0.02in}
  \renewcommand\tabcolsep{1.0pt}
  \renewcommand\arraystretch{1.0}
  \scalebox{1.0}{
  \begin{tabular}{|p{1.59cm}<{\centering} || p{1.58cm}<{\centering} | p{1.68cm}<{\centering} | p{1.68cm}<{\centering} | p{1.68cm}<{\centering} | p{1.68cm}<{\centering} | p{1.68cm}<{\centering} | p{1.78cm}<{\centering} | p{1.78cm}<{\centering} | p{1.78cm}<{\centering} | p{1.78cm}<{\centering} | p{1.68cm}<{\centering} |}
  \hlineB{2.2}
  \rowcolor{lightgray}
  Dataset& Metric  & CBDNet~\cite{guo2019toward} & MPRNet~\cite{mehri2021mprnet} & MIRNet~\cite{zamir2020learning} & NBNet~\cite{cheng2021nbnet} & GMSNet~\cite{song2020grouped} & Uformer~\cite{wang2021uformer} &  HINet~\cite{chen2021hinet}  & BDE \\ \hline \hline
  \multirow{2}{*}{SIDD~\cite{SIDD_2018_CVPR}}
  & PSNR  & 30.78 & 39.71 & 39.72 & 39.76 &  39.47 &  39.77 & 39.99   & \bf{40.04}  \\ 
  & SSIM  & 0.801 & 0.958 & 0.959 &  0.966 &  0.968 &  0.970 & 0.970  & \bf{0.971}  \\ 
    \hlineB{2.2}
  \end{tabular}
  }
  \vspace{-0.04in}
  \label{tab:sidd}
\end{table*}

\subsection{Implement details}~\label{sec:imp_detail}
For real-world image denoising, DND~\cite{plotz2017benchmarking}  and SIDD~\cite{SIDD_2018_CVPR} are the most commonly used benchmark.
DND contains 50 pairs of real-world noisy images and corresponding ground truth images for evaluation.
However, it doesn't provide any noisy/clean training pairs for training, and the ground truth images for evaluation are not publicly available.
The PSNR and SSIM values can be obtained by the online server after submitting the noise removal results, and only 10 samples with size $512\times512$ are provided the detailed quantitative index. 
On the contrary, SIDD consists of both training pairs, validation pairs, and testing pairs.
Here, we follow \cite{song2020grouped} to use the medium version of SIDD (320 image pairs) as the real-world training set, and the validation pairs are utilized as our testing set.
Note that the images of both DND and SIDD are more than 4K resolution.
To synthesize the noisy/clean image pairs, we employ the real noise model in CBDNet~\cite{guo2019toward} with DIV2K dataset~\cite{Agustsson_2017_CVPR_Workshops} to generate the huge number of training pairs with cropped size $256 \times 256$.
The noise generating parameters are the same as CBDNet.

To train the scoring network, we employ the Adam optimizer with $\beta_1=0.9$ and $\beta_2=0.999$.
The mini-batch size is set to 32, and $128\times128$ patch size is randomly cropped from the dataset during training.
The default data augmentations, such as flipping, rotating, and cropping, for better generalization.
Note that Mix-Up~\cite{wang2021uformer} is also adopted during the training stage.
The initial learning rate is set to 1e-4 and is decayed with factor 0.5 after every 200 epochs until 1e-6.
Our scoring network is trained by 2000 epochs on one NVIDIA RTX 2080Ti GPU with two days.
For each epoch, it only contains 100 iterations and is totally trained for 2$\times$10$^5$ iterations.
The code is built on the PyTorch platform and is released at \url{https://github.com/lpj0/BDE}.

\subsection{Comparison with State-of-the-arts}
\subsubsection{Evaluation on Real-world Noisy Images}
For evaluating the effectiveness of our proposed method, we first published the results on DND and SIDD benchmarks.
We collect 10 famous denoisers, \emph{i.e.} BM3D~\cite{dabov2007image},  DnCNN~\cite{zhang2017beyond} 
CBDNet~\cite{guo2019toward}, CycleISP~\cite{zamir2020cycleisp},  MPRNet~\cite{mehri2021mprnet},  MIRNet~\cite{zamir2020learning},  NBNet~\cite{cheng2021nbnet},  GMSNet~\cite{song2020grouped},  Uformer~\cite{wang2021uformer}, and HINet~\cite{chen2021hinet} for comparison.
Note that our method should first fuse different denoisers, \emph{i.e.} taking the denoised results of different methods as input.
To fuse different characteristic denoisers, we first classify the denoisers into three types in terms of training sets:
i) only using the medium version of SIDD, \emph{e.g.} NBNet, HINet, and Uformer;
ii) using both the synthetic dataset and the medium version of SIDD, \emph{e.g.} GMSNet, CycleISP.
iii) mainly using the synthetic dataset, but also with other negligible real-world datasets, \emph{e.g.} CBDNet.
And the fused denoisers should contain all types to make their respective advantages complementary to each other. 
Then in terms of the PSNR value and class type, we select five denoisers, \emph{i.e.} CBDNet, NBNet, HINet, Uformer, and GMSNet, as our denoiser pool.

\noindent
{\bf{Results on DND dataset.}} 
As shown in Table.~\ref{tab:dnd}, our proposed method outperforms the state-of-the-art GMSNet 0.28 dB.
Among the fused denoisers, GMSNet owns the highest performance in terms of PSNR and SSIM metrics, which is trained on both the synthetic noisy/clean pairs and the medium version of SIDD.  
In comparison with CBDNet, our method outperforms their 2.4dB. 
Even if the mean PSNR/SSIM performance of other fused denoisers, \emph{i.e.} CBDNet,  MPRNet,  MIRNet,  NBNet, and Uformer, are much lower than GMSNet, they can still provide useful information for reconstructing the restored image,  which indicates the effectiveness of our proposed method.
We also provide visual results for comparison, as shown in Fig.~\ref{fig:dndv}.
The hot maps of weights and uncertainties are also provided for verifying the effectiveness of the proposed method.
One can find that different parts of the denoised results are selected to combine.
Then the final denoised image can be further improved. 

We also trained a big model named ``ModelH" that combines the fused denoisers and increases convolution channels to match the same Flops and parameters of the overall model.
As listed in Tab.~\ref{tab:dnd}, ModelH obtains similar PSNR to BDE on SSID but far lower on DND, which demonstrates the performance gain is not got by the increased Flops or parameters but by the fused denoisers. 
 We provide running time and PSNR indexes in Tab.~\ref{tab:factor}.
 More than 96$\%$ of the running time is spent on the inference of the fused denoisers, while our scoring network is fast enough for predicting weighting maps.
 
\noindent
{\bf{Results on SIDD dataset.}} 
We also evaluate our method on SIDD as listed in Tab.~\ref{tab:sidd}.
Note that SIDD has the corresponding training set, and most methods are trained on it to get higher PSNR/SSIM performance.
Even though, our method also can get a higher PSNR value than the previous state-of-the-art method.

\subsubsection{Comparison with CSNet}
As mentioned above, CSNet~\cite{choi2019optimal} is also an ensemble method that can combine several denoisers by estimating MSE and convex optimization.
However, it is only training and testing on gray-style Gaussian image denoising.
To fair comparison, we also apply our ensemble method to gray-style Gaussian image denoising.
The setting is kept the same with CSNet:
REDNet~\cite{mao2016image} is initial denoiser.
Five well-trained denoisers, \emph{i.e.} $\{ \mathbf{D}_{G}^{10},\mathbf{D}_{G}^{20},\mathbf{D}_{G}^{30},\mathbf{D}_{G}^{40},\mathbf{D}_{G}^{50}\}$ with noise level $\{10,20,30,40,50\}$, are token as the denoiser pool $\mathbb{D}$.
And the noise is not clipped in our experiment.
Note that we only compare the combination part, \emph{i.e.} CSNet before booster (denoted as CSNet-B).
Table.~\ref{tab:bsd200} lists the PSNR results of CSNet-B and our ensemble method on BSD68 dataset~\cite{martin2001database}.
One can find that our ensemble method BDE can achieve similar or better results than CSNet-B.

\begin{table}[!t]\footnotesize
  \centering
  \caption{
  Quantitative results of PSNR (dB) on BSD68 dataset for gray-style Gaussian denoising. $\mathbf{D}_{G}^{10}$ denotes REDNet which is trained with noise level 10, and NL stands for the noise level. One can find our proposed method owns the compared method with CSNet-B.
  {\bf{Bold}} font indicates the best performance.
  }
  \vspace{-0.0in}
  \renewcommand\tabcolsep{1.0pt}
  \renewcommand\arraystretch{1.2}
  \scalebox{1.0}{
  \begin{tabular}{|p{0.69cm}<{\centering} || p{0.88cm}<{\centering} | p{0.88cm}<{\centering} | p{0.88cm}<{\centering} | p{0.88cm}<{\centering} | p{0.88cm}<{\centering} | p{1.62cm}<{\centering} | p{1.08cm}<{\centering} | p{1.08cm}<{\centering} | p{1.08cm}<{\centering} | p{1.08cm}<{\centering} | p{1.58cm}<{\centering} |}
  \hlineB{2.2}
  \rowcolor{lightgray}
  NL  & $\mathbf{D}_{G}^{10}$ & $\mathbf{D}_{G}^{20}$ & $\mathbf{D}_{G}^{30}$ & $\mathbf{D}_{G}^{40}$ & $\mathbf{D}_{G}^{50}$ & CSNet-B~\cite{choi2019optimal} & BDE \\ \hline \hline
  10 &  \bf{34.17} & 30.75  &  28.25   & 27.03 & 25.97 & 34.14 & \bf{34.17}   \\ 
  \rowcolor{tinygray} 15 &  28.25 & 30.89  &  28.33   & 27.08 & 25.99 & 31.46 &   \bf{31.47}       \\
  20 &  24.19 & \bf{30.48}  &  28.46   & 27.15 & 26.03 & \bf{30.48} & \bf{30.48}       \\
  \rowcolor{tinygray} 25 &  21.68 & 26.65  &  28.61   & 27.24 & 26.08 & \bf{29.07} &   \bf{ 29.07}       \\
  30 &  19.86 & 22.91  &  \bf{28.52}   & 27.35 & 26.14 & \bf{28.52} & \bf{28.52}       \\
  \rowcolor{tinygray} 35 &  18.43 & 20.52  &  26.56   & 27.44 & 26.28 & 27.72 &     \bf{27.73}     \\
  40 &  17.25 & 18.84  &  23.23   & \bf{27.24} & 26.33 & \bf{27.24} & \bf{27.24}       \\
  \rowcolor{tinygray} 45 &  16.25 & 17.54  &  20.77   & 25.37 & 26.41 & 26.66 &  \bf{26.67}        \\
  50 &  15.38 & 16.45  &  18.95   & 22.51 & \bf{26.32} & \bf{26.32} & \bf{26.32}       \\
    \hlineB{2.2}
  \end{tabular}
  }
  \vspace{-0.0in}
  \label{tab:bsd200}
\end{table}
\begin{table}[!t]\footnotesize
  \centering
  \caption{
    {
  Comparison of CSNet-B and our proposed method on DND benchmark.
  {\bf{Bold}} font indicates the best performance.}
  }
  \vspace{-0.0in}
  \renewcommand\tabcolsep{1.0pt}
  \renewcommand\arraystretch{1.2}
  \scalebox{1.0}{
  \begin{tabular}{|p{1.3cm}<{\centering} || p{1.42cm}<{\centering} | p{1.82cm}<{\centering}| p{1.82cm}<{\centering} | p{1.82cm}<{\centering} | p{1.68cm}<{\centering} |}
  \hlineB{2.2}
  \rowcolor{lightgray}
  Dataset& Metric & GMSNet~\cite{song2020grouped} & CSNet-B~\cite{choi2019optimal}  & BDE  \\ \hline \hline
     \multirow{2}{*}{DND~\cite{plotz2017benchmarking}} 
  & PSNR & 40.24 & 40.25 &  \bf{40.52}     \\
  & SSIM & 0.962 &  0.962 & \bf{0.964}     \\
    \hlineB{1.5}
  \end{tabular}
  }
  \vspace{-0.00in}
  \label{tab:csnetdnd}
\end{table}

\noindent
{\bf{Evaluating on real-world DND benchmark.}}
We further retrain CSNet-B with the same training set as our BDE in Sec.\ref{sec:imp_detail} and evaluate it on the real-world DND benchmark for a fair comparison.
As shown in Table.~\ref{tab:csnetdnd}, CSNet-B is only comparable with GMSNet, and far less than our BDE.
We assume that estimating MSE is uncertain when it comes across an unknown benchmark.
While our method can obtain stable results by introducing DBL to the scoring network.

\subsubsection{Fusing Gaussian Denoisers for Real-world Noisy Images}
We train five color Gaussian denoisers as our denoiser pool and test our ensemble method on the DND benchmark to verify the effect of processing the unknown real-world noisy image.
The architecture of those denoisers is CDnCNN~\cite{zhang2017beyond}, which is a well-known denoiser that can be fast fine-tuned with the different noise levels.
We denote those denoisers as $\{ \mathbf{D}_{C}^{nl}\}$ with noise level $nl=\{10,20,30,40,50\}$.
We directly utilize the proposed method to combine Gaussian denoisers, denoted as BDE$_{GD}$.
Then the fused result with its input noisy image is taken as training pairs for fine-tuning the best denoisers, which can be seen as knowledge distillation with self-information.
We name this variant as BDE$_{GD}$+D.

Table.~\ref{tab:combinegaussian} lists the results on the DND benchmark. 
Note that CDnCNN-B is not reported by its authors, and we get this result from \cite{zhou2020awgn}.
Even though combining several Gaussian denoiser, it can easily outperform CBDNet, which utilizes the complex noise synthesis pipeline to simulate real-world noise.
This indicates our proposed method owns good generalization ability.
Furthermore, after knowledge distillation, the performance of our method can be further improved.
Therefore, we assume that the unknown type of noisy image can be handled by our ensemble method.
\begin{table}[!h]\footnotesize
\vspace{0.08in}
  \centering
  \caption{
The PSNR/SSIM values of combining only Gaussian denoisers on the DND benchmark. 
    BDE$_{GD}$ directly combines five Gaussian denoisers, while BDE$_{GD}$+D utilizes knowledge distillation to fine-tune the best Gaussian denoisers by taking the input noisy images and the fused results as training pairs.
    {\bf{Bold}} font indicates the best performance.
  }
  
  \renewcommand\tabcolsep{1.0pt}
  \renewcommand\arraystretch{1.2}
  \scalebox{0.94}{
  \begin{tabular}{|p{1.3cm}<{\centering} || p{0.82cm}<{\centering} | p{1.95cm}<{\centering}| p{1.62cm}<{\centering} | p{1.45cm}<{\centering} | p{1.55cm}<{\centering} |}
  \hlineB{2.2}
  \rowcolor{lightgray}
  Dataset& Metric& CDnCNN-B~\cite{zhang2017beyond} & CBDNet~\cite{guo2019toward}  & BDE$_{GD}$  &  BDE$_{GD}$+D  \\ \hline \hline
\multirow{2}{*}{DND~\cite{plotz2017benchmarking}}
  & PSNR & 32.43 & 38.06 & 38.30 &  \bf{38.60}     \\
  & SSIM & 0.790 & 0.942 & 0.945 &  \bf{0.947}     \\
    \hlineB{1.5}
  \end{tabular}
  }
  \vspace{-0.0in}
  \label{tab:combinegaussian}
\end{table}

\begin{figure*}[!htbp]
    \vspace{-0ex}
    \centering
      \begin{minipage}[c]{0.24\textwidth}
      \vspace{0.5ex}
      \centering
        \centerline{\includegraphics[width=0.99\linewidth]{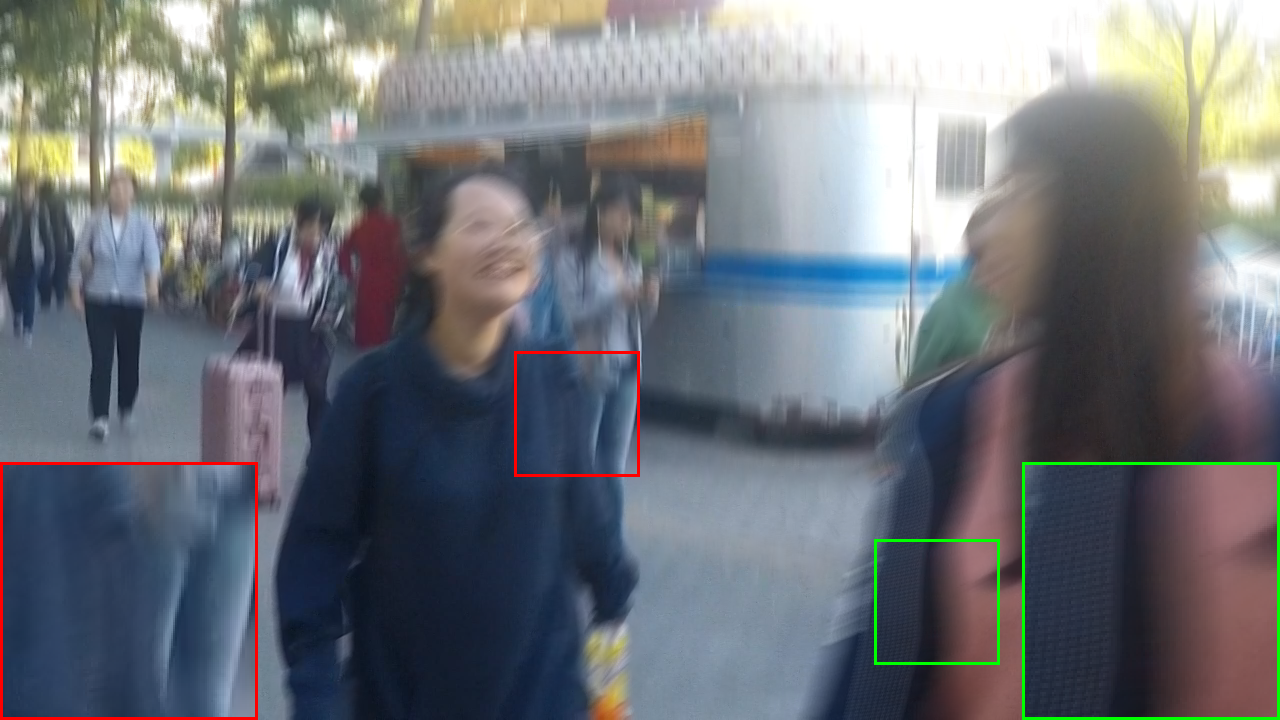}}
        \centerline{Blur Image}
        \label{fig:Blur_hide}
      \end{minipage}%
    \hspace{0.ex}
      \begin{minipage}[c]{0.24\textwidth}
      \centering
        \centerline{\includegraphics[width=0.99\linewidth]{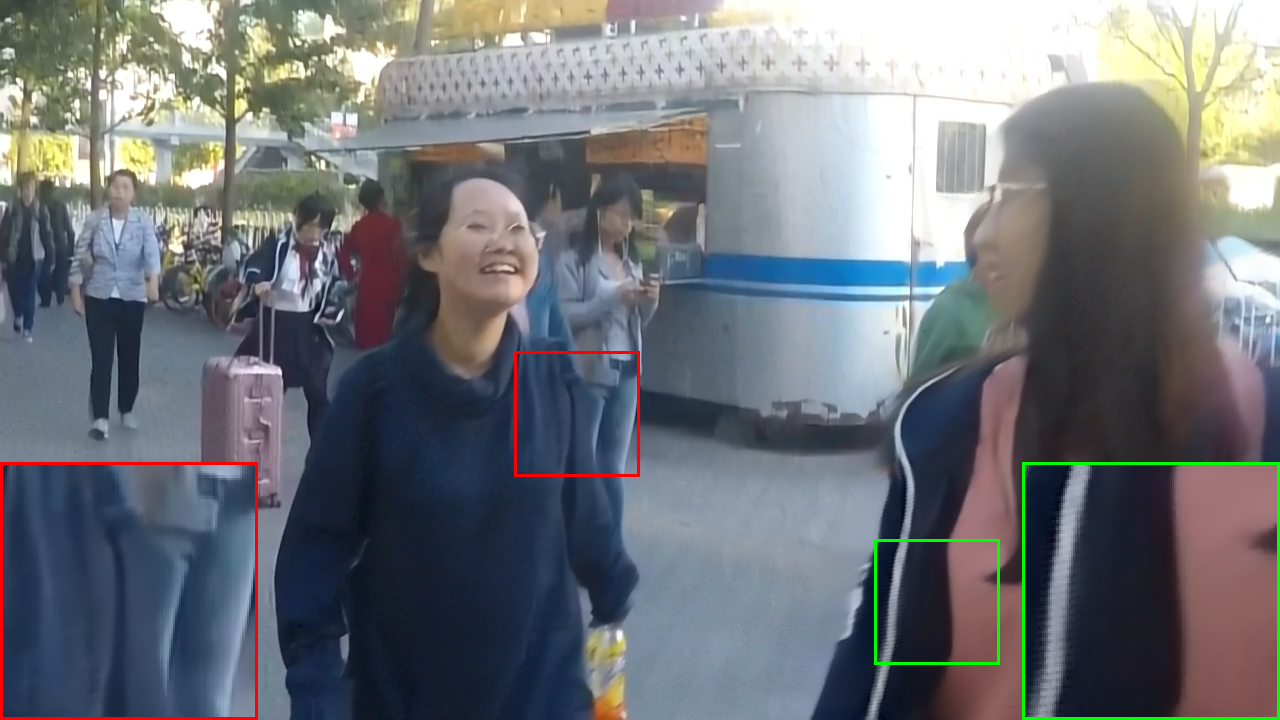}}
        \centerline{MTRNN~\cite{park2020multi}}
        \label{fig:MTRNN_hide}
      \end{minipage}%
    \hspace{-0.ex}
      \begin{minipage}[c]{0.24\textwidth}
      \centering
        \centerline{\includegraphics[width=0.99\linewidth]{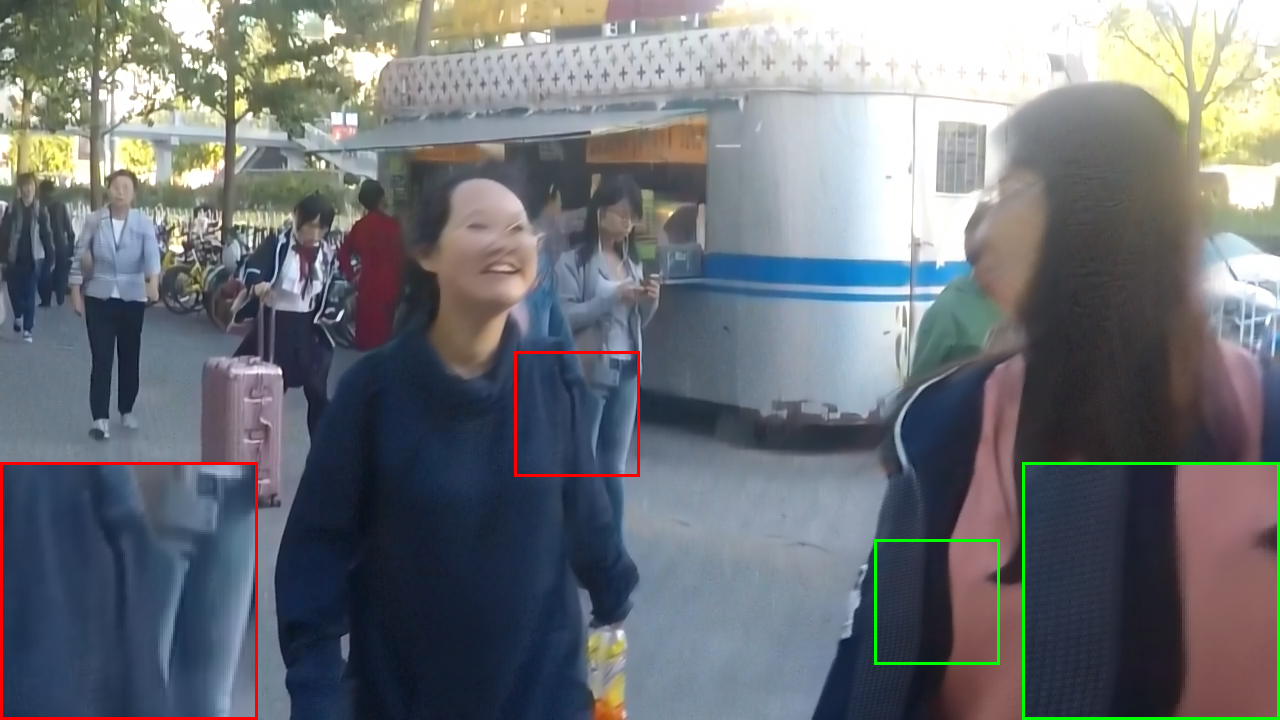}}
        \centerline{MPRNet~\cite{mehri2021mprnet}}
        \label{fig:MPRNet_hide}
      \end{minipage}%
    \hspace{-0.ex}
      \begin{minipage}[c]{0.24\textwidth}
      \centering
        \centerline{\includegraphics[width=0.99\linewidth]{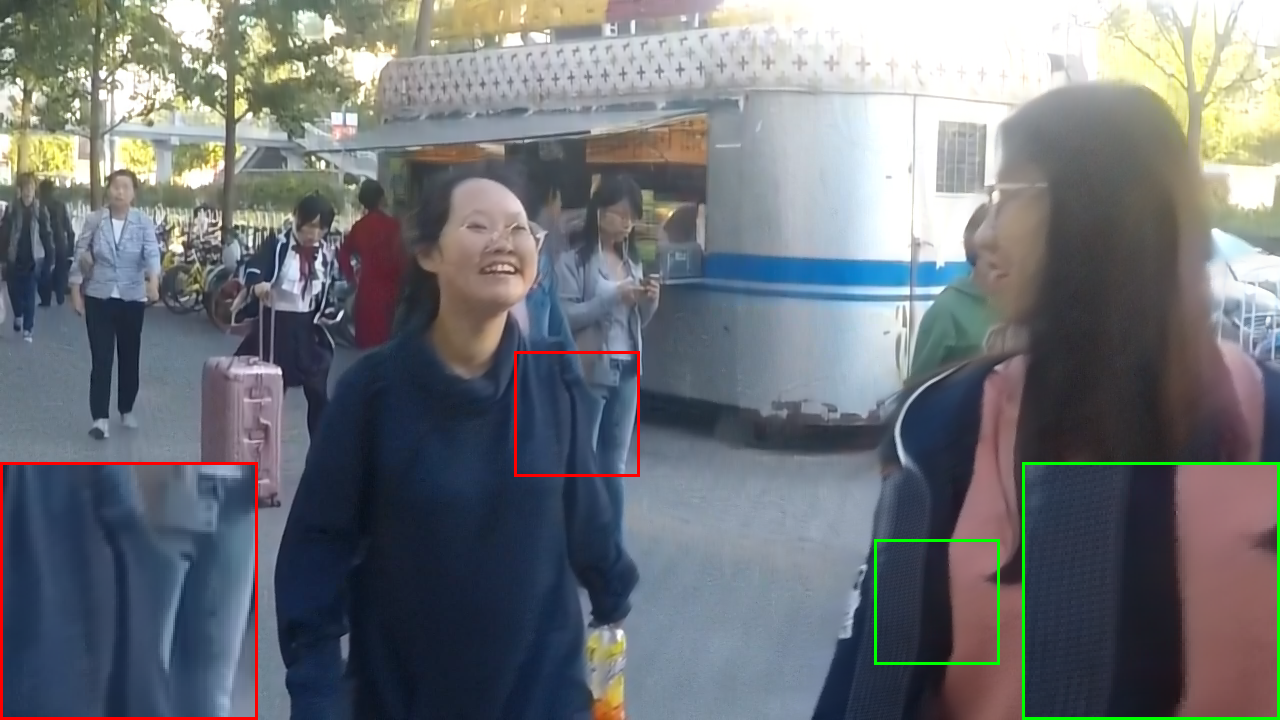}}
        \centerline{MIMO-UNet+~\cite{cho2021rethinking}}
        \label{fig:mimo_hide}
      \end{minipage}%
      
    \hspace{0.ex}
      \begin{minipage}[c]{0.24\textwidth}
      \centering
        \centerline{\includegraphics[width=0.99\linewidth]{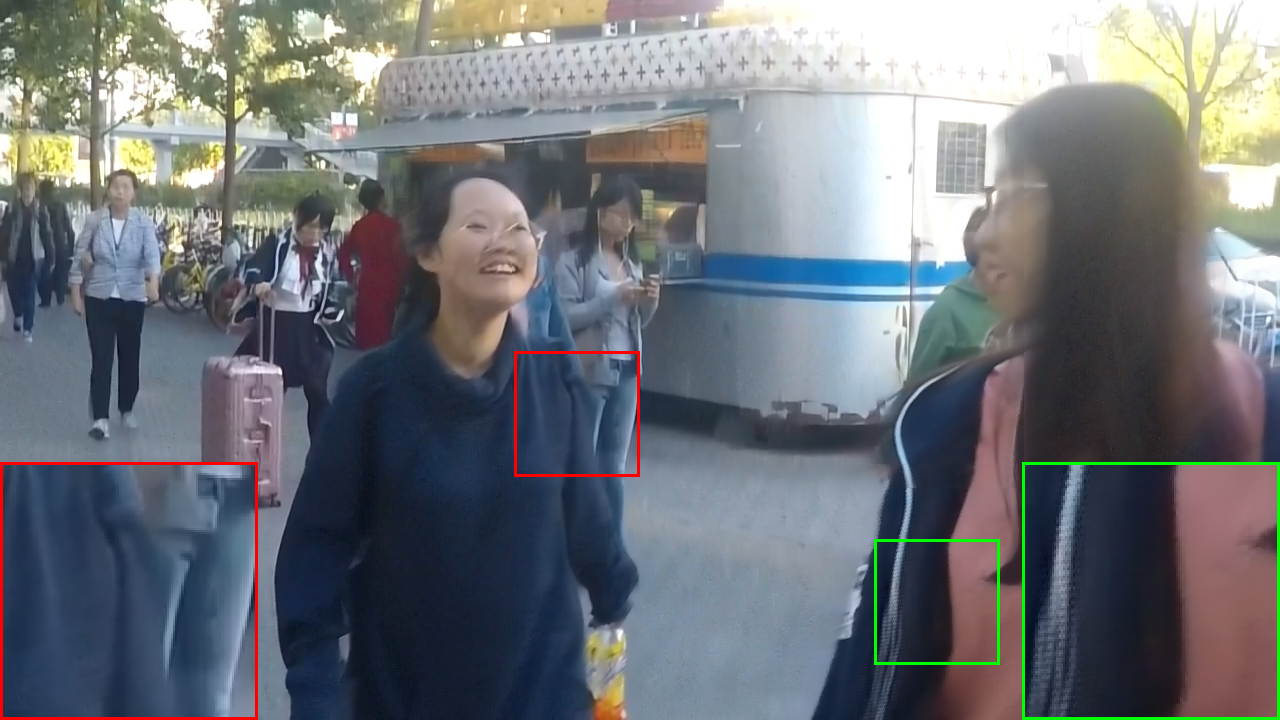}}
        \centerline{HINet~\cite{chen2021hinet}}
        \label{fig:hinet_hide}
      \end{minipage}%
    \hspace{-0.ex}
      \begin{minipage}[c]{0.24\textwidth}
      \centering
        \centerline{\includegraphics[width=0.99\linewidth]{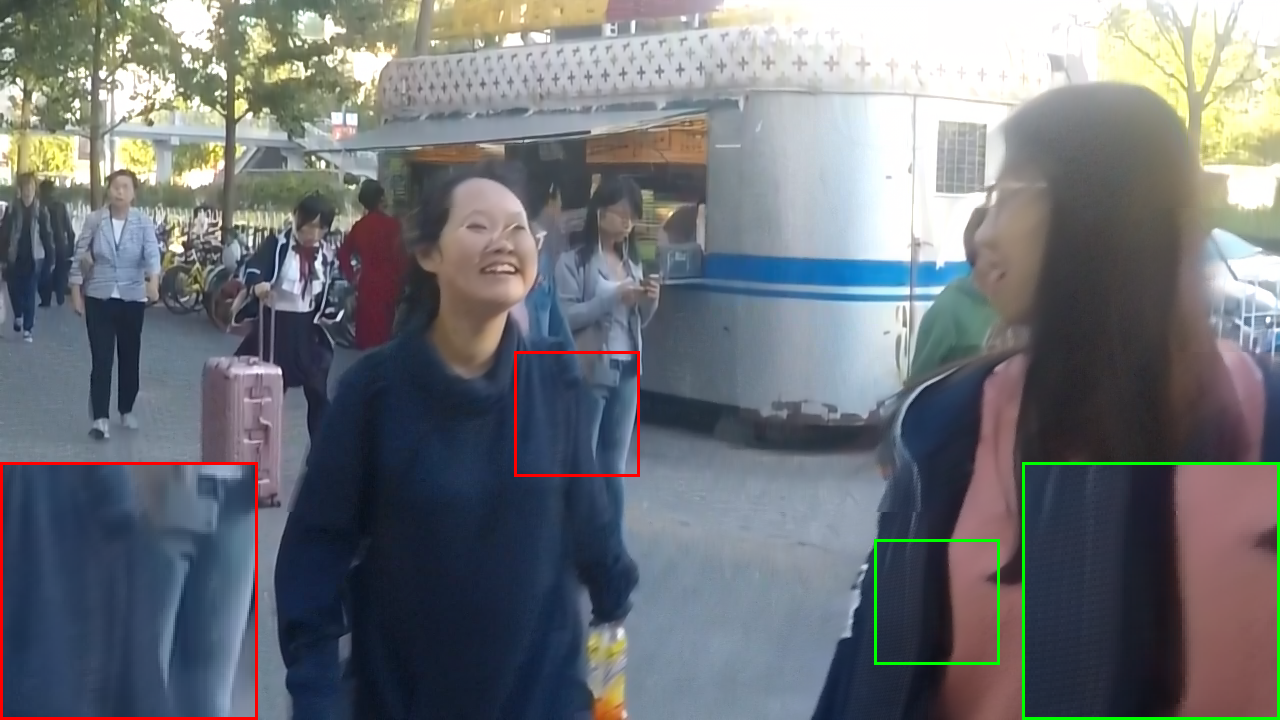}}
        \centerline{Restormer~\cite{Zamir2021Restormer}}
        \label{fig:Restormer_hide}
      \end{minipage}%
    \hspace{-0.ex}
      \begin{minipage}[c]{0.24\textwidth}
      \centering
        \centerline{\includegraphics[width=0.99\linewidth]{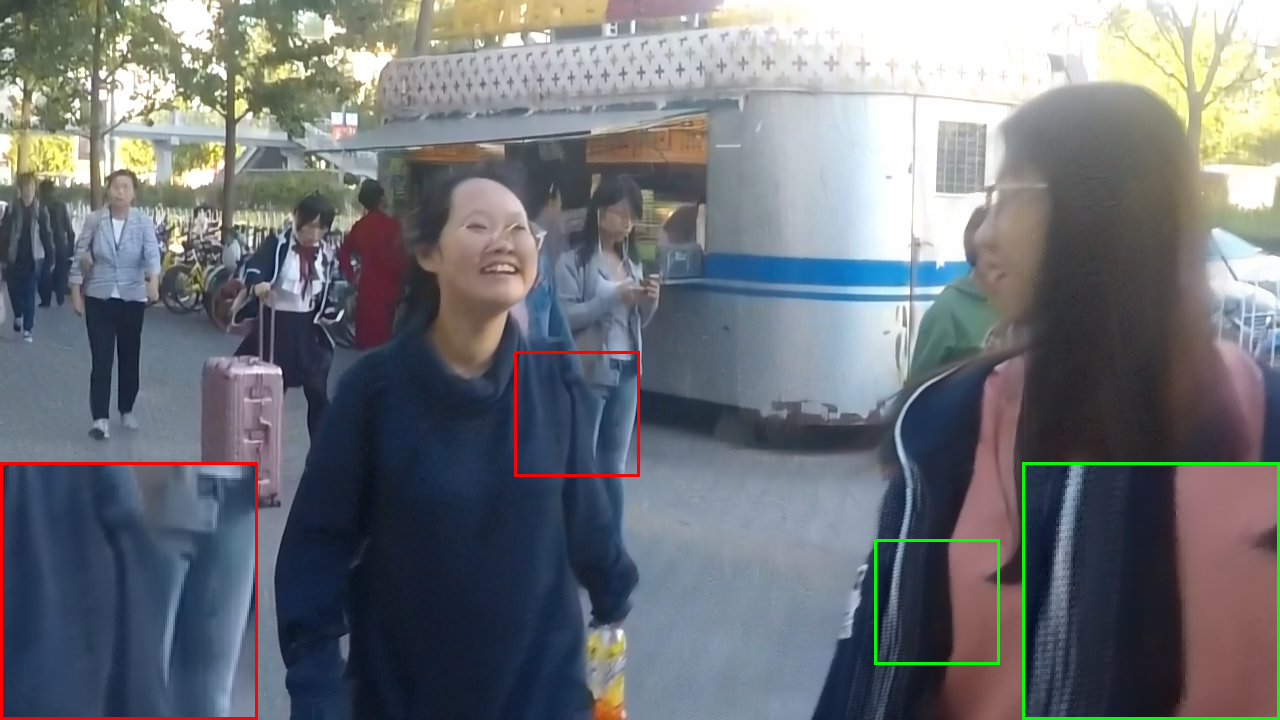}}
        \centerline{BDE}
        \label{fig:hide_u2em}
      \end{minipage}%
     \hspace{-0.ex}
    \begin{minipage}[c]{0.24\textwidth}
      \centering
        \centerline{\includegraphics[width=0.99\linewidth]{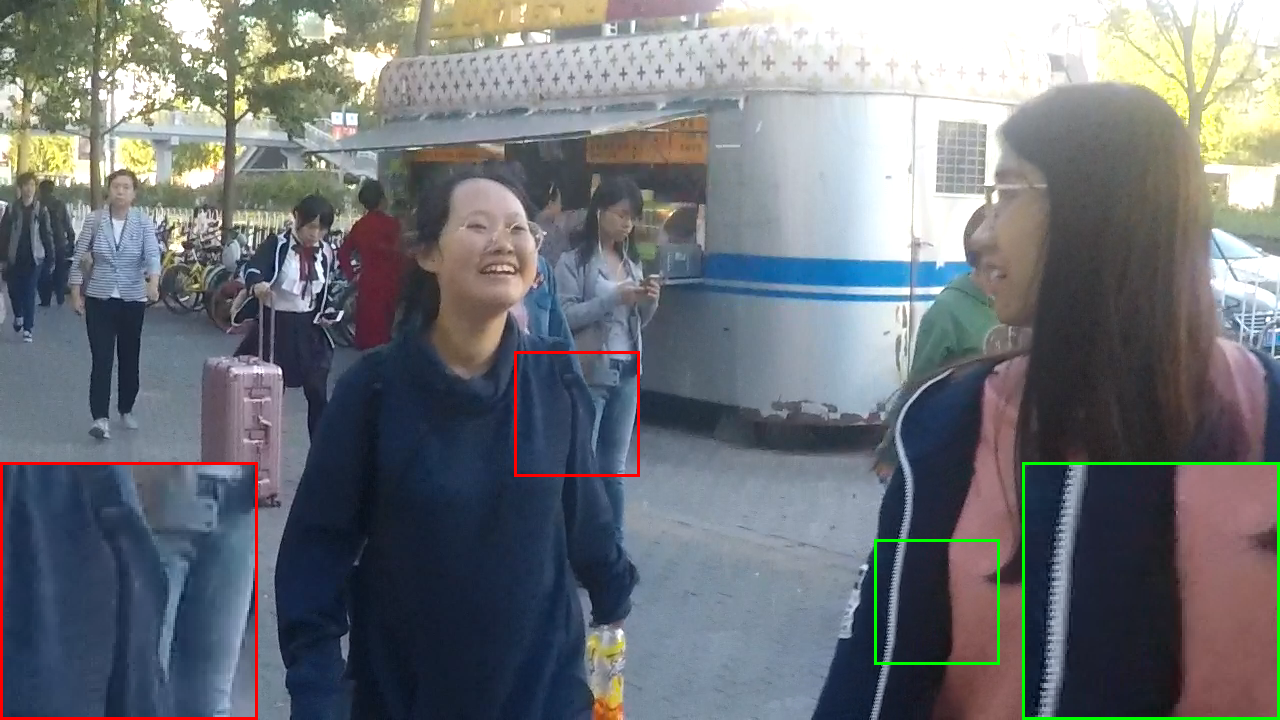}}
        \centerline{Ground Truth}
        \label{fig:hide_gt}
      \end{minipage}
    \vspace{-1.ex}
      \caption{ The visual results ``93fromGOPR1089.MP4" of different deblurring methods on HIDE.
      For the region in the red box, our BDE is composed of linearly combining the results of two methods, \emph{i.e.}  MTRNN~\cite{park2020multi} and MIMO-UNet+~\cite{cho2021rethinking}.
      While in the red box, our BDE mainly comes from HINet~\cite{chen2021hinet}.
      }
      \label{fig:hide}
      \vspace{-0.0ex}
    \end{figure*}
  
\begin{figure*}[!htbp]
    \vspace{-2ex}
    \centering
      \begin{minipage}[c]{0.12\textwidth}
      \vspace{0.5ex}
      \centering
        \centerline{\includegraphics[width=0.99\linewidth]{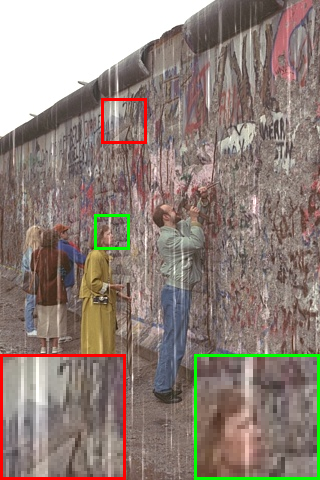}}
        \centerline{Rainy Image}
        \label{fig:rainy_rain100l_16}
      \end{minipage}%
    \hspace{-0.2ex}
      \begin{minipage}[c]{0.12\textwidth}
      \centering
        \centerline{\includegraphics[width=0.99\linewidth]{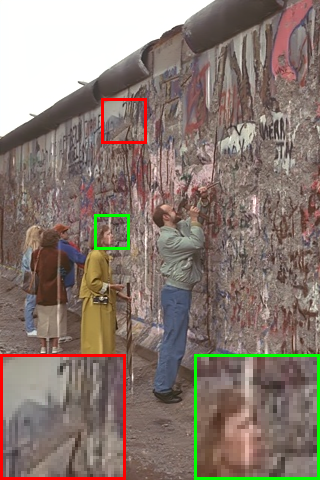}}
        \centerline{PreNet~\cite{ren2019progressive}}
        \label{fig:MTRNN_rain100l_16}
      \end{minipage}%
    \hspace{-0.2ex}
      \begin{minipage}[c]{0.12\textwidth}
      \centering
        \centerline{\includegraphics[width=0.99\linewidth]{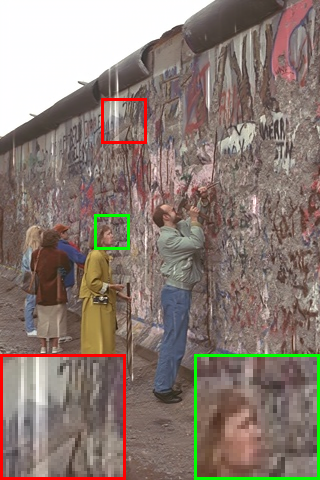}}
        \centerline{MSPFN~\cite{mehri2021mprnet}}
        \label{fig:MPRNet_rain100l_16}
      \end{minipage}%
    \hspace{-0.2ex}
      \begin{minipage}[c]{0.12\textwidth}
      \centering
        \centerline{\includegraphics[width=0.99\linewidth]{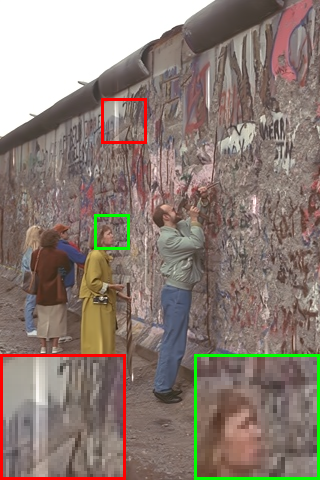}}
        \centerline{MPRNet~\cite{mehri2021mprnet}}
        \label{fig:mimo_rain100l_16}
      \end{minipage}%
    \hspace{-0.2ex}
      \begin{minipage}[c]{0.12\textwidth}
      \centering
        \centerline{\includegraphics[width=0.99\linewidth]{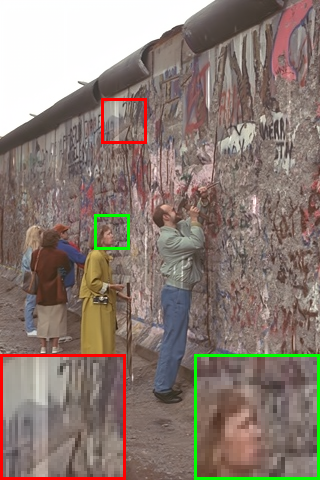}}
        \centerline{HINet~\cite{chen2021hinet}}
        \label{fig:hinet_rain100l_16}
      \end{minipage}%
    \hspace{-0.2ex}
      \begin{minipage}[c]{0.12\textwidth}
      \centering
        \centerline{\includegraphics[width=0.99\linewidth]{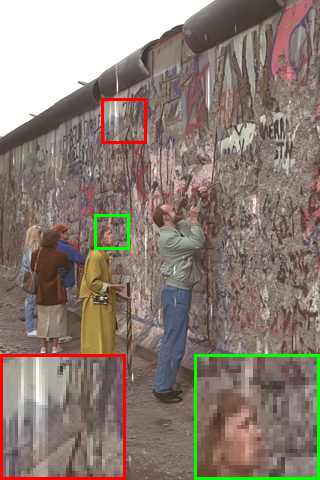}}
        \centerline{Restormer~\cite{Zamir2021Restormer}}
        \label{fig:Restormer_rain100l_16}
      \end{minipage}%
    \hspace{-0.2ex}
      \begin{minipage}[c]{0.12\textwidth}
      \centering
        \centerline{\includegraphics[width=0.99\linewidth]{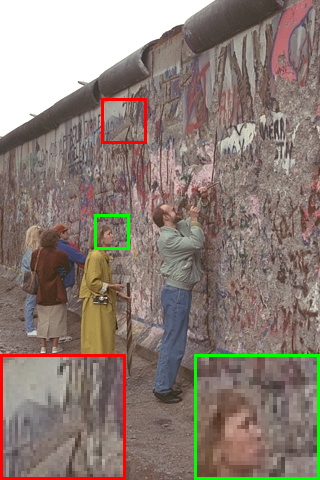}}
        \centerline{BDE}
        \label{fig:rain100l_16_u2em}
      \end{minipage}%
     \hspace{-0.2ex}
    \begin{minipage}[c]{0.12\textwidth}
      \centering
        \centerline{\includegraphics[width=0.99\linewidth]{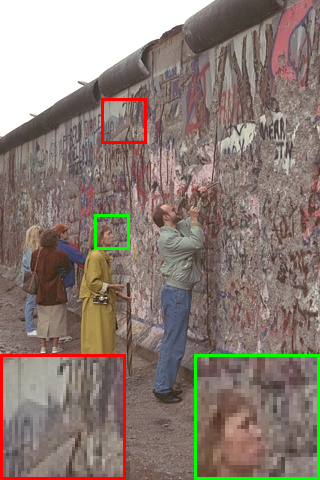}}
        \centerline{Ground Truth}
        \label{fig:rain100l_16_gt}
      \end{minipage}
    \vspace{-1.ex}
      \caption{ The visual results are the image ``100'' of different deraining methods on the Rain100L benchmark.
      For our result, the region in the red box is mainly selected from PreNet~\cite{ren2019progressive}, while the region in the green box is mainly selected from MPRNet~\cite{mehri2021mprnet}.
      }
      \label{fig:rain100l_16}
      \vspace{-0.0ex}
\end{figure*}

\begin{table*}[!htbp]\footnotesize
\centering
\vspace{-0.04in}
\caption{
Comparison with different models for image deblurring. 
Image deblurring is tested on GoPro~\cite{nah2017deep}, HIDE~\cite{shen2019human}, RealBlur-R~\cite{rim2020real} and RealBlur-J~\cite{rim2020real}.
}
\vspace{-0.02in}
\renewcommand\tabcolsep{1.0pt}
\renewcommand\arraystretch{1.2}
\scalebox{0.999}{
\begin{tabular}{|p{1.9cm}<{\centering} || p{1.88cm}<{\centering} | p{2.38cm}<{\centering} | p{1.68cm}<{\centering} | p{1.88cm}<{\centering} | p{2.28cm}<{\centering} | p{1.68cm}<{\centering}| p{1.68cm}<{\centering}| p{1.68cm}<{\centering}|}
\hlineB{2.99}
\rowcolor{lightgray} PSNR$\uparrow$/SSIM$\uparrow$ & DeblurGAN~\cite{kupyn2018deblurgan}& DeblurGAN-v2~\cite{Kupyn_2019_ICCV} & MTRNN~\cite{park2020multi} & MPRNet~\cite{mehri2021mprnet} & MIMO-UNet+~\cite{cho2021rethinking} & HINet~\cite{chen2021hinet} & Restormer~\cite{Zamir2021Restormer} & BDE \\ \hline \hline 
GoPro~\cite{nah2017deep}  &  28.70 / 0.858   &  29.55 / 0.934  &  31.15 / 0.945  &  32.66 / 0.959  &  32.45 / 0.957   & 32.71 / 0.959  &  32.92 / 0.961   &   \bf{33.38} /  \bf{0.963}    \\
\rowcolor{tinygray} HIDE~\cite{shen2019human} &  24.51 / 0.871    &  26.61 / 0.875   & 29.15 / 0.918 &  30.96 / 0.939 & 29.99 / 0.930  & 30.33 / 0.932 &  31.22 / 0.942   & \bf{31.52} / \bf{0.946}      \\
RealBlur-R~\cite{rim2020real}& 33.79  / 0.903 &   35.26  / 0.944  &  35.79 / 0.951  &   35.99  / 0.952  &  35.54 / 0.947  &  33.80 / 0.947   &  36.19 / 0.957    & \bf{36.44} / \bf{0.959}      \\
\rowcolor{tinygray} RealBlur-J~\cite{rim2020real} & 27.97 / 0.834 &  28.70 / 0.866 & 28.44  / 0.862  &  28.70 / 0.873 &  27.63 / 0.837  & 26.32 / 0.854   &   28.96 /  0.879  & \bf{29.14} / \bf{0.883}      \\ \hline
Average &   28.74 / 0.867 &  30.03 / 0.905 &  31.13 / 0.919  &   32.08 / 0.931 &  31.40 / 0.918  & 30.79 / 0.922  &  32.32 / 0.935    & \bf{32.62} / \bf{0.937}      \\
\hlineB{1.5}
\end{tabular}
}
\label{tab:gopro}
\vspace{-0.05in}
\end{table*}

\begin{table*}[!htbp]\footnotesize
\centering
\vspace{-0.00in}
\caption{
Comparison with different models for image deraining. 
Image deraining is tested on five well-known datasets, \emph{i.e.} Test100~\cite{zhang2019image},  Rain100H~\cite{yang2017deep}, Rain100L~\cite{yang2017deep}, Test2800~\cite{fu2017removing}, and Test1200~\cite{zhang2018density}.
}
\vspace{-0.02in}
\renewcommand\tabcolsep{1.0pt}
\renewcommand\arraystretch{1.2}
\scalebox{0.999}{
\begin{tabular}{|p{1.9cm}<{\centering} || p{1.88cm}<{\centering} | p{1.88cm}<{\centering} | p{1.88cm}<{\centering} | p{1.88cm}<{\centering} | p{1.88cm}<{\centering} |  p{1.88cm}<{\centering} | p{1.88cm}<{\centering} | p{1.88cm}<{\centering}|}
\hlineB{2.99}
\rowcolor{lightgray} PSNR$\uparrow$/SSIM$\uparrow$ & DerainNet~\cite{fu2017clearing} & SPAIR~\cite{purohit2021spatially} & PreNet~\cite{ren2019progressive} & MSPFN~\cite{Kui_2020_CVPR} & MPRNet~\cite{mehri2021mprnet} & HINet~\cite{chen2021hinet} & Restormer~\cite{Zamir2021Restormer} & BDE \\ \hline \hline 
Test100~\cite{zhang2019image}  & 22.77 / 0.810 & 30.35 / 0.909 & 24.81 / 0.851 & 27.50 / 0.876 & 30.27 / 0.897 & 30.29 / 0.906  & 32.00 / 0.923 & \bf{32.07} / \bf{0.925}   \\
\rowcolor{tinygray} Rain100H~\cite{yang2017deep} & 14.92 / 0.592 & 30.95 / 0.892 & 26.77 / 0.858 & 28.66 / 0.860 & 30.41 / 0.890 & 30.65 / 0.894  & 31.46 / 0.904 & \bf{31.58} / \bf{0.905}   \\
Rain100L~\cite{yang2017deep} & 27.03 / 0.884 & 36.93 / 0.969 & 32.44 / 0.950 & 32.40 / 0.933 & 36.40 / 0.965 & 37.28 / 0.970  & 38.99 / 0.978 & \bf{39.38} / \bf{0.983}   \\
\rowcolor{tinygray}  Test2800~\cite{fu2017removing} & 24.31 / 0.861 & 33.34 / 0.936 & 31.75 / 0.916 & 32.82 / 0.930 & 33.64 / 0.938 & 33.91 / 0.941  & 34.18 / 0.944 & \bf{34.30} / \bf{0.944}   \\
Test1200~\cite{zhang2018density} & 23.38 / 0.835 & 33.04 / 0.922 & 31.36 / 0.911 & 32.39 / 0.916 & 32.91 / 0.916 & 33.05 / 0.919 & 33.19 / 0.926 & \bf{33.37} / 0.925  \\ \hline
\rowcolor{tinygray} Average   & 22.48 / 0.796 & 32.91 / 0.926 &  29.42 / 0.897  & 30.75 / 0.903  &  32.73 / 0.921  &  33.03 / 0.926 &   33.96 / 0.935 &    \bf{34.14} / \bf{0.942}   \\
\hlineB{1.5}
\end{tabular}
}
\label{tab:deraining}
\vspace{-0.05in}
\end{table*}

\begin{table*}[!htbp]\footnotesize
\centering
\vspace{-0.00in}
\caption{
Comparison with different models for single image super-resolution on scale factor $\times4$. 
SISR is tested on five well-known datasets, \emph{i.e.} Set5~\cite{bevilacqua2012low},  Set14~\cite{zeyde2010single}, BSD100~\cite{martin2001database}, Urban100~\cite{huang2015single}, and Manga109~\cite{matsui2017sketch}.
}
\vspace{-0.02in}
\renewcommand\tabcolsep{1.0pt}
\renewcommand\arraystretch{1.2}
\scalebox{0.999}{
\begin{tabular}{|p{1.9cm}<{\centering} || p{1.88cm}<{\centering} | p{1.88cm}<{\centering} | p{1.88cm}<{\centering} | p{1.88cm}<{\centering} | p{1.88cm}<{\centering} |  p{1.88cm}<{\centering} | p{1.88cm}<{\centering} | p{1.88cm}<{\centering}|}
\hlineB{2.99}
\rowcolor{lightgray} PSNR$\uparrow$/SSIM$\uparrow$ & EDSR~\cite{lim2017enhanced} & RNAN~\cite{zhang2019rnan} & RCAN~\cite{zhang2020residual} & SAN~\cite{dai2019second} & RRDB~\cite{wang2018esrgan} & IPT~\cite{chen2021pre} & SwinIR~\cite{liang2021swinir} & BDE \\ \hline \hline 
Set5~\cite{bevilacqua2012low}     & 32.46 / 0.897 & 32.49 / 0.898 & 32.63 / 0.900 & 32.64 / 0.900 & 32.73 / 0.901 & 32.64 / 0.899 & 32.92 / 0.904  & \bf{33.00} / \bf{0.905}    \\
\rowcolor{tinygray} Set14~\cite{zeyde2010single}    & 28.80 / 0.788 & 28.83 / 0.788 & 28.87 / 0.789 & 28.92 / 0.789 & 28.99 / 0.792 & 29.01 / 0.793 & 29.09 / 0.795  & \bf{29.13} / \bf{0.796}     \\
BSD100~\cite{martin2001database}   & 27.71 / 0.742 & 27.72 / 0.742 & 27.77 / 0.744 & 27.78 / 0.744 & 27.85 / 0.746 & 27.82 / 0.748 & 27.92 / 0.749  & \bf{27.95} / \bf{0.750}     \\
\rowcolor{tinygray} Urban100~\cite{huang2015single} & 26.64 / 0.803 & 26.61 / 0.802 & 26.82 / 0.809 & 26.79 / 0.807 & 27.03 / 0.815 & 27.26 / 0.821 & 27.45 / 0.825  & \bf{27.63} / \bf{0.831}   \\
Manga109~\cite{matsui2017sketch} & 31.02 / 0.915 & 31.09 / 0.915 & 31.22 / 0.917 & 31.18 / 0.917 & 31.66 / 0.920 & 31.62 / 0.922 & 32.03 / 0.926  & \bf{32.30} / \bf{0.930}    \\ \hline
\rowcolor{tinygray} Average   & 29.34 / 0.829 & 29.36 / 0.829 & 29.46 / 0.832  & 29.46 / 0.832 & 29.65 / 0.835  & 29.67 / 0.837 & 29.88 / 0.840  & \bf{30.00} / \bf{0.842}     \\

\hlineB{1.5}
\end{tabular}
}
\label{tab:sisr}
\vspace{-0.00in}
\end{table*}

\begin{table*}[!ht]\footnotesize
  \centering
  \caption{
  The PSNR/SSIM values of different variants on the DND benchmark.
  Compared with BDE$_{Normal}$, BDE(-MC), BDE$\left(-\boldsymbol{\sigma}\right)$, BDE($_{pl}$),  BDE$_{Weight}$, and BDE(-${\ell_{sum}}$), our proposed method can obtain more than 0.1dB gain.
  By using both negative log-likelihood loss $\ell_{nll}$ and fusing loss $\ell_{fuse}$, the final method surpass GMSNet 0.28dB.
  }
  \vspace{-0.0in}
  \renewcommand\tabcolsep{1.0pt}
  \renewcommand\arraystretch{1.2}
  \scalebox{1.0}{
  \begin{tabular}{|p{1.3cm}<{\centering} || p{1.2cm}<{\centering} | p{1.72cm}<{\centering}| p{1.52cm}<{\centering} | p{1.52cm}<{\centering} | p{1.52cm}<{\centering} | p{1.62cm}<{\centering}| p{1.72cm}<{\centering} | p{1.72cm}<{\centering} | p{1.62cm}<{\centering} | p{1.52cm}<{\centering} | p{1.52cm}<{\centering} | p{1.52cm}<{\centering}|}
  \hlineB{2.2}
  \rowcolor{lightgray}
  Dataset& Metric& GMSNet~\cite{song2020grouped} & BDE$_{Normal}$ & BDE(-MC) & BDE$\left(-\boldsymbol{\sigma}\right)$ & BDE($_{PaW}$)  &  BDE$_{Weight}$   & BDE(-${\ell_{fuse}}$)   & BDE(-${\ell_{nll}}$) & BDE(Final)  \\ \hline \hline
   \multirow{2}{*}{DND~\cite{plotz2017benchmarking}}
  & PSNR  & 40.24  & 40.31 & 40.35 & 40.34 & 40.29 &  40.33  & 40.46 & 40.47 &\bf{40.52}  \\
  & SSIM & 0.962   & 0.962 & 0.962 & 0.963 &  0.962 &  0.962 &  0.963 &  0.963 &  \bf{0.964}     \\
    \hlineB{1.5}
  \end{tabular}
  }
  \vspace{-0.0in}
  \label{tab:vardnd}
\end{table*}

\subsection{Extension to Other Image Restoration Tasks}\label{sec:experiments extension}
Our proposed BDE method can also be generalized to other image restoration tasks, \emph{e.g.} image deblurring, image deraining, and single image super-resolution.
Similar to image denoising, we also select five representative models as the model pool for training and testing.
Other training settings can also refer to Sec.~\ref{sec:imp_detail}.

\subsubsection{Image deblurring.}
As listed in Tab.~\ref{tab:gopro}, seven state-of-the-art methods, \emph{i.e.}
DeblurGAN~\cite{kupyn2018deblurgan}, DeblurGAN-v2~\cite{Kupyn_2019_ICCV}, MTRNN~\cite{park2020multi}, MPRNet~\cite{mehri2021mprnet}, MIMO-UNet+~\cite{cho2021rethinking} , HINet~\cite{chen2021hinet}, and Restormer~\cite{Zamir2021Restormer}, are selected for comparison.
And MTRNN~\cite{park2020multi}, MPRNet~\cite{mehri2021mprnet}, MIMO-UNet+~\cite{cho2021rethinking}, HINet~\cite{chen2021hinet}, and Restormer~\cite{Zamir2021Restormer} are utilized as fused models.
It can easily find that our method obtains $\sim$0.4dB gain on image deblurring over Restormer~\cite{Zamir2021Restormer}.
We also provide the visual results as shown in Fig.~\ref{fig:hide}.
Our method can combine the advantages of the different fused methods, which leads to the best results.

\subsubsection{Image deraining.}
We also select seven image deraining models, \emph{i.e.} DerainNet~\cite{fu2017clearing}, SPAIR~\cite{purohit2021spatially}, PreNet~\cite{ren2019progressive}, MSPFN~\cite{Kui_2020_CVPR},  MPRNet~\cite{mehri2021mprnet}, HINet~\cite{chen2021hinet}, and Restormer~\cite{Zamir2021Restormer}, for comparison.
And PreNet~\cite{ren2019progressive}, MSPFN~\cite{Kui_2020_CVPR},  MPRNet~\cite{mehri2021mprnet}, HINet~\cite{chen2021hinet}, and Restormer~\cite{Zamir2021Restormer} is taken as the deraining pool.
As listed in Tab.~\ref{tab:deraining}, our BDE can outperform the state-of-the-art method, \emph{i.e.} Restormer, $>0.3$dB.
Fig.~\ref{fig:rain100l_16} is visual results of different comparison methods. 
It indicates that our method can well fuse different datasets with a variety of rain streaks.

\subsubsection{Single image super-resolution}
We select five state-of-the-art SISR methods, \emph{i.e.}
RCAN~\cite{zhang2020residual}, SAN~\cite{dai2019second}, RRDB~\cite{wang2018esrgan}, IPT~\cite{chen2021pre}, and SwinIR~\cite{liang2021swinir}, as the SR pool.
And EDSR~\cite{lim2017enhanced} and RNAN~\cite{zhang2019rnan} are also selected for comparison.
As listed in Tab.~\ref{tab:sisr}, compared to the state-of-the-art SwinIR model, the PSNR can obtain $\sim$0.1dB gain.
Hence, our proposed method can be applied to more low-level tasks.
Employing a similar strategy for the task would give a more tangible method.

\subsection{Ablation study}
We design two experiments to notarize the role of different parts in our proposed method:
i) why and how to estimate scoring map;
ii) influence on the type and number of fused denoisers.

\noindent
{\bf{Different approaches to fuse denoisers.}} 
As mentioned in Sec.~\ref{ensemble_model}, it exists several approaches for fusing different models in image restoration tasks.
We implement those approaches, \emph{i.e.}  \cite{liao2015video,liu2017robust,cho2021compression,yang2020image}, by ourselves due to no public code or different tasks.
\cite{liu2020improved} is not reproduced due to only fusing two denoisers and lack of implementation details to fuse more denoisers. 
The same denoiser pool with our BDE is utilized for a fair comparison.
Note that \cite{yang2020image} stacks the denoisers in sequence.
As listed in Tab.~\ref{tab:DND_c}, our BDE can easily outperform the other fusion strategies which verify the effectiveness of the proposed method.
The reason why they are far inferior to our method is that they focused on fusing the models under the same condition, \emph{e.g.}  same noise type, noise level, and even same architecture.
Moreover, they all take fixed methods as input, which limits their application to real scenarios.
\begin{table}[!ht]\footnotesize
\centering
\vspace{-0.0in}
\caption{
Comparison with different fusion strategies.
We state that \cite{liao2015video}, \cite{liu2017robust}, \cite{cho2021compression}, and \cite{yang2020image} are implemented by ourselves because of no public code or different tasks.
They also fuse the same denoisers with our BDE.
Note that \cite{yang2020image} stacks the denoisers and is retrained by the same training dataset.
}
\vspace{-0.0in}
\renewcommand\tabcolsep{1.0pt}
\renewcommand\arraystretch{1.2}
\scalebox{0.99}{
\begin{tabular}{|p{1.24cm}<{\centering} | p{1.12cm}<{\centering} | p{1.12cm}<{\centering} | p{1.12cm}<{\centering} | p{1.12cm}<{\centering} | p{1.12cm}<{\centering} | p{1.12cm}<{\centering} | p{0.96cm}<{\centering} | p{1.12cm}<{\centering} |}
\hlineB{2.0}
\rowcolor{lightgray} Dataset & Metric & \cite{liao2015video} & \cite{liu2017robust} & \cite{cho2021compression} & \cite{yang2020image}   & BDE \\ \hline \hline
\specialrule{0em}{0.5pt}{.5pt}
\multirow{2}{*}{DND~\cite{plotz2017benchmarking}} 
& PSNR    &  40.26   &  40.28  &  40.31  &  40.35 &  \bf{40.52}    \\
& SSIM    &  0.960   &  0.961  &  0.960  &  0.961 &  \bf{0.964} \\  
\specialrule{0em}{-0.5pt}{-.5pt}
\hlineB{1.5}
\end{tabular}
}
\label{tab:DND_c}
\vspace{-0.0in}
\end{table}


\noindent
{\bf{Effect of Bayesian deep learning.}}
We examine the necessity of Bayesian deep learning (BDL) in our BDE by training three variants:
1) BDE$_{Normal}$ is a normal U-Net architecture without MC dropout layers and estimating uncertainty map $\boldsymbol{\sigma}$. 
2) BDE(-MC) removes MC dropout layers. 
3) BDE$\left(-\boldsymbol{\sigma}\right)$ doesn't estimate uncertainty map $\boldsymbol{\sigma}$.
One can find the generalization ability of the model becomes weaker without BDL.
Both MC dropout layers and estimating uncertainty map $\boldsymbol{\sigma}$ improve the denoising performance.

\noindent
{\bf{The effectiveness of estimating scoring map.}}
We utilize three variants of the scoring network to verify the effectiveness of estimating the scoring map: 
1) BDE($_{PaW}$) denotes patch-level prediction instead of pixel-wise prediction;
2) BDE$_{Weight}$ denotes directly estimating the weighting map without the SoftMax function.
Table.~\ref{tab:vardnd} lists the results of different variants, and we can obtain the following  conclusions:
i) Instead of predicting a single score for one patch, the pixel-wise prediction can be more effective to combine different denoisers' results.
ii) Because of the SoftMax function, our prediction can be more robust. 

\noindent
{\bf{The role of the different loss functions.}} 
In Eq.(\ref{eq:weights}), our loss function contain two terms, \emph{i.e.}  negative log-likelihood loss in Eq.(\ref{eq:nlltv}) and  fusing loss in Eq.(\ref{eq:sum}).
We remove one of them and denote them as BDE(-${\ell_{nll}}$) and BDE(-${\ell_{fuse}}$) respectively.
As shown in Table.~\ref{tab:vardnd}, the performance of the proposed method is obtained by their joint effort.

\begin{figure}[!t]
  \vspace{-0ex}
  \begin{center}
    \vspace{-0.0ex}
    \includegraphics[width=0.49\textwidth]{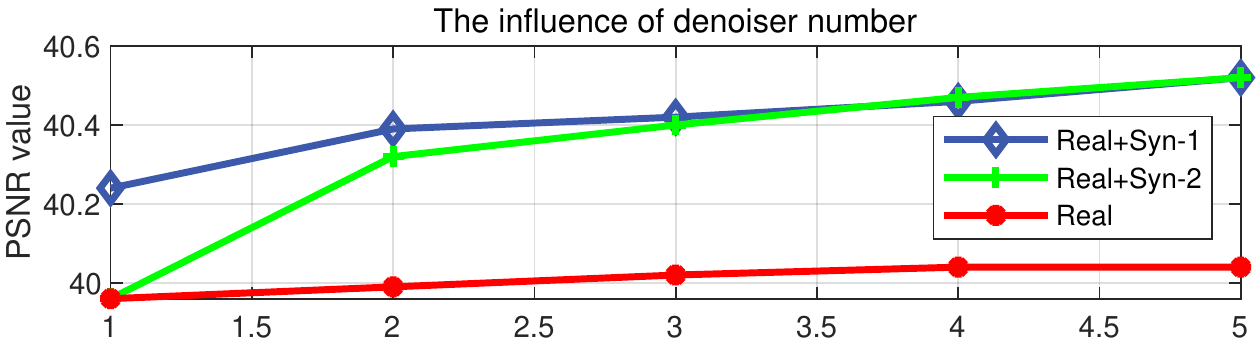}
    \vspace{-0.0ex}
  \end{center}
    \caption{The influence of the fused denoisers' number. 
    The red line denotes that only fuses the real-world denoisers, and the base denoiser is Uformer. 
    While results of both the green line and the blue line are that fuse both real-world and synthetic denoisers. 
    The base denoiser of the green line is Uformer, and the base denoiser of the blue line is GMSNet.}
    \label{fig:num_denoisers}
    \vspace{-0.0ex}
\end{figure}

\noindent
{\bf{The number of fused denoisers.}}
Fig.~\ref{fig:num_denoisers} shows the influence of the fused denoisers' number.
The red line denotes that only fuses the real-world denoisers.  
The base denoiser is Uformer~\cite{wang2021uformer}, and then NBNet~\cite{cheng2021nbnet}, MIRNet~\cite{zamir2020learning}, MPRNet~\cite{mehri2021mprnet}, and HINet are added in turn. 
Both the green line and the blue line that fuse both real-world and synthetic denoisers.
The base denoiser of the green line is Uformer~\cite{wang2021uformer}, and then CBDNet~\cite{guo2019toward},  NBNet~\cite{cheng2021nbnet}, MPRNet~\cite{mehri2021mprnet}, HINet~\cite{cheng2021nbnet}, and GMSNet~\cite{song2020grouped} are added in turn.
The base denoiser of the blue line is GMSNet~\cite{song2020grouped}, and then CBDNet~\cite{guo2019toward},  NBNet~\cite{cheng2021nbnet}, MIRNet~\cite{zamir2020learning}, MPRNet~\cite{mehri2021mprnet}, and HINet~\cite{cheng2021nbnet} are added in turn.
One can find that the PSNR performance can get a continuous rise with the number of denoisers increasing, but the growth rate slows down after fusing 3 denoisers.
Compared with the red and green lines, the PSNR performance can obtain a remarkable gain after combining the denoisers with different training datasets.
Compared with the red and blue lines, the increase rate of the red line is lower due to all denoisers overfitting on the SIDD dataset.
So we suggest that the fused denoisers should own different characteristics, especially training on different datasets.

We provide the results of fusing four of five different denoisers in Tab.~\ref{tab:number}.
Then we can get the following conclusions:
i) fusing denoisers with different training datasets can achieve better generalization than one dataset;
ii) greater difference of network, stronger complementary;
iii) collecting denoisers with different training datasets and architectures rather than fusing more denoisers.   
\begin{table}[!hp]\footnotesize
\centering
\vspace{-0.0in}
\caption{
The PSNR  of fusing four denoisers by our BDE. 
Here we label the denoisers as:
a) GMSNet~\cite{song2020grouped}, b) CBDNet~\cite{guo2019toward}, c) Uformer~\cite{wang2021uformer}, d) HINet~\cite{chen2021hinet}, and e) NBNet~\cite{cheng2021nbnet}.
}
\vspace{-0.00in}
\renewcommand\tabcolsep{1.0pt}
\renewcommand\arraystretch{1.2}
\scalebox{0.99}{
\begin{tabular}{|p{1.28cm}<{\centering} |p{0.9cm}<{\centering} |p{0.96cm}<{\centering}| p{0.96cm}<{\centering} | p{0.96cm}<{\centering} | p{0.96cm}<{\centering} | p{0.96cm}<{\centering}| p{0.96cm}<{\centering}| p{0.96cm}<{\centering}|}
\hlineB{2.2}
\rowcolor{lightgray}  Dataset & Metric & abcd & abce &  abde  & acde & bcde & abcde \\ \hline \hline 
\multirow{2}{*}{DND~\cite{plotz2017benchmarking}} 
& PSNR   & 40.47  &  40.46 &  40.41  & 40.34    & 40.41 &   \bf{40.52}   \\
& SIMM   & 0.962  &  0.963 & 0.961   &  0.960   & 0.963 &   \bf{0.964}   \\
\hlineB{1.5}
\end{tabular}
}
\label{tab:number}
\vspace{-0.00in}
\end{table}

\noindent
{\bf{Complementarity  of the different denoisers' results}}
The complementarity  of the different denoisers' results comes from two folds:
i) The training datasets. 
Assume that two denoisers with the same architecture are trained on two extremely different datasets,  \emph{e.g.} real noisy dataset and synthetic dataset, respectively. 
As shown in Fig.~\ref{fig:num_denoisers}, only fusing denoisers that only train on SIDD cannot achieve big progress on the DND dataset. 
While the performance can achieve big gain by fusing denoisers training on different datasets.
ii) The architecture. 
For example, Uformer~\cite{wang2021uformer} considers more global information, while GMSNet~\cite{song2020grouped} pays more attention to local information.

\subsection{Limitations}
The limitations of our proposed method can be summarized in two aspects:
On the one hand, the fused denoisers should have different characteristics to form complementarity.
Then the denoising performance can be further improved.
If not so, the performance is limited, such as results on SIDD.
On the other hand, our method is time-consuming due to inference time on different denoisers and estimating weighting maps for each denoised result, even though it only needs 0.002 seconds to estimate a scoring map with size $256\times256$ .
This limits our method of applying to real-time scenes. 
However, for real-time scenarios, BDE has been trained to generalize well compared to single small models, and therefore we leverage it as an oracle to get noise-free results from the unpaired inputs, and subsequently, any practical small model designed for real-time inference can be trained in a supervised manner.

\section{Conclusion}\label{secV}
In this paper, we propose an effective Bayesian deep ensemble (BDE) method to exploit different external prior, \emph{i.e.} different well-trained denoisers, for real-world image denoising and other image restoration tasks.
Our BDE takes one of the noise removal results with its noisy image as input and then predicts the scoring map.
After obtaining all scoring maps, we use the SoftMax function to obtain the probability maps to combine the images in a pixel-wise approach. 
Without the corresponding training dataset, our BDE method can fuse different denoisers to achieve state-of-the-art performance on DND dataset, which indicates the robustness and generalization of our proposed ensemble method.
Moreover, by combining five Gaussian denoisers, it can obtain better denoising performance than CBDNet.
Hence, for an unknown type of noisy image, our BDE method is a potentially effective approach to provide a clean result.
For other image restoration tasks, our BDE method can be directly applied and also achieve an amazing effect.  
However, we also point out the limitations of our work, such as the special demands of different denoisers and time-consuming.
In further work, we will improve those limitations and seek an unsupervised approach for ensemble learning. 

\ifCLASSOPTIONcaptionsoff
  \newpage
\fi

\bibliographystyle{IEEEtran}
\bibliography{egbib}

\end{document}